\pdfoutput=1

\documentclass[11pt]{article}

\usepackage[final]{ACL2023}

\usepackage{fancyvrb}

\usepackage{subcaption}
\usepackage{booktabs}
\usepackage{graphicx}
\usepackage{multirow}

\usepackage{times}
\usepackage{latexsym}
\usepackage{array}
\usepackage{siunitx}

\usepackage[T1]{fontenc}

\usepackage[utf8]{inputenc}

\usepackage{microtype}

\usepackage{inconsolata}

\usepackage{graphicx}
\usepackage{multicol}

\usepackage{xcolor}

\usepackage{import}

\usepackage{colortbl}
\usepackage{xcolor}
\usepackage{tabu}

\usepackage{amsmath,amsfonts,amssymb}
\usepackage{algorithm}
\usepackage{algpseudocode}
\graphicspath{{figures/}}

\makeindex

\definecolor{tblue}{RGB}{93, 142, 150}
\definecolor{tred}{RGB}{191, 97, 106}
\definecolor{dlblue}{RGB}{216, 235, 255}
\definecolor{dgreen}{RGB}{124, 155, 127}
\definecolor{dpink}{RGB}{207, 166, 208}
\definecolor{dyellow}{RGB}{255, 248, 199}
\definecolor{dgray}{RGB}{46, 49, 49}

\newcommand{\durl}[1]{\textcolor{tblue}{\underline{\url{#1}}}}

\newcounter{DaveDefCounter}
\title{Tokenization Matters: Navigating Data-Scarce Tokenization for Gender Inclusive Language Technologies}

\author{%
   \textbf{Anaelia Ovalle\textsuperscript{\rm 1}\thanks{~ This work was done when Anaelia Ovalle was an intern at Amazon. Correspondence to anaelia@cs.ucla.edu and mninareh@amazon.com.} 
  \quad Ninareh Mehrabi\textsuperscript{\rm 2} 
  \quad Palash Goyal\textsuperscript{\rm 2}} 
  \\ 
  \textbf{Jwala Dhamala\textsuperscript{\rm 2} 
  \quad Kai-Wei Chang\textsuperscript{\rm 2} 
  \quad Richard Zemel\textsuperscript{\rm 2} 
  \quad Aram Galstyan\textsuperscript{\rm 2}} 
  \\ 
  \textbf{Yuval Pinter\textsuperscript{\rm 3 \rm 2}\thanks{~ Yuval Pinter holds concurrent appointments as a Senior Lecturer of CS at Ben-Gurion University and as an Amazon Visiting Academic. This paper describes work performed at Ben-Gurion University and is not associated with his role at Amazon.}  
  \quad Rahul Gupta\textsuperscript{\rm 2}}\\
\textsuperscript{\rm 1}University of California, Los Angeles \\ 
\textsuperscript{\rm 2}Amazon AGI Foundations \\
\textsuperscript{\rm 3}Ben-Gurion University of the Negev, Be'er Sheva, Israel \\ 
}

\begin{document}

\maketitle

\begin{abstract}
Gender-inclusive NLP research has documented the harmful limitations of gender binary-centric large language models (LLM), such as the inability to correctly use gender-diverse English neopronouns (e.g., xe, zir, fae). 
While data scarcity is a known culprit, the precise mechanisms through which scarcity affects this behavior remain underexplored.
We discover LLM misgendering is significantly influenced by Byte-Pair Encoding (BPE) tokenization, the tokenizer powering many popular LLMs.
Unlike binary pronouns, BPE overfragments neopronouns, a direct consequence of data scarcity during tokenizer training.
This disparate tokenization mirrors tokenizer limitations observed in multilingual and low-resource NLP, unlocking new misgendering mitigation strategies.
We propose two techniques: (1) \textit{pronoun tokenization parity}, a method to enforce consistent tokenization across gendered pronouns, and (2) utilizing pre-existing LLM pronoun knowledge to improve neopronoun proficiency. Our proposed methods outperform finetuning with standard BPE, improving neopronoun accuracy from 14.1\% to 58.4\%. 
Our paper is the first to link LLM misgendering to tokenization and deficient neopronoun grammar, indicating that LLMs unable to correctly treat neopronouns as pronouns are more prone to misgender.

\end{abstract}

\begin{figure}[!t]
    \centering
    \includegraphics[width=\linewidth]{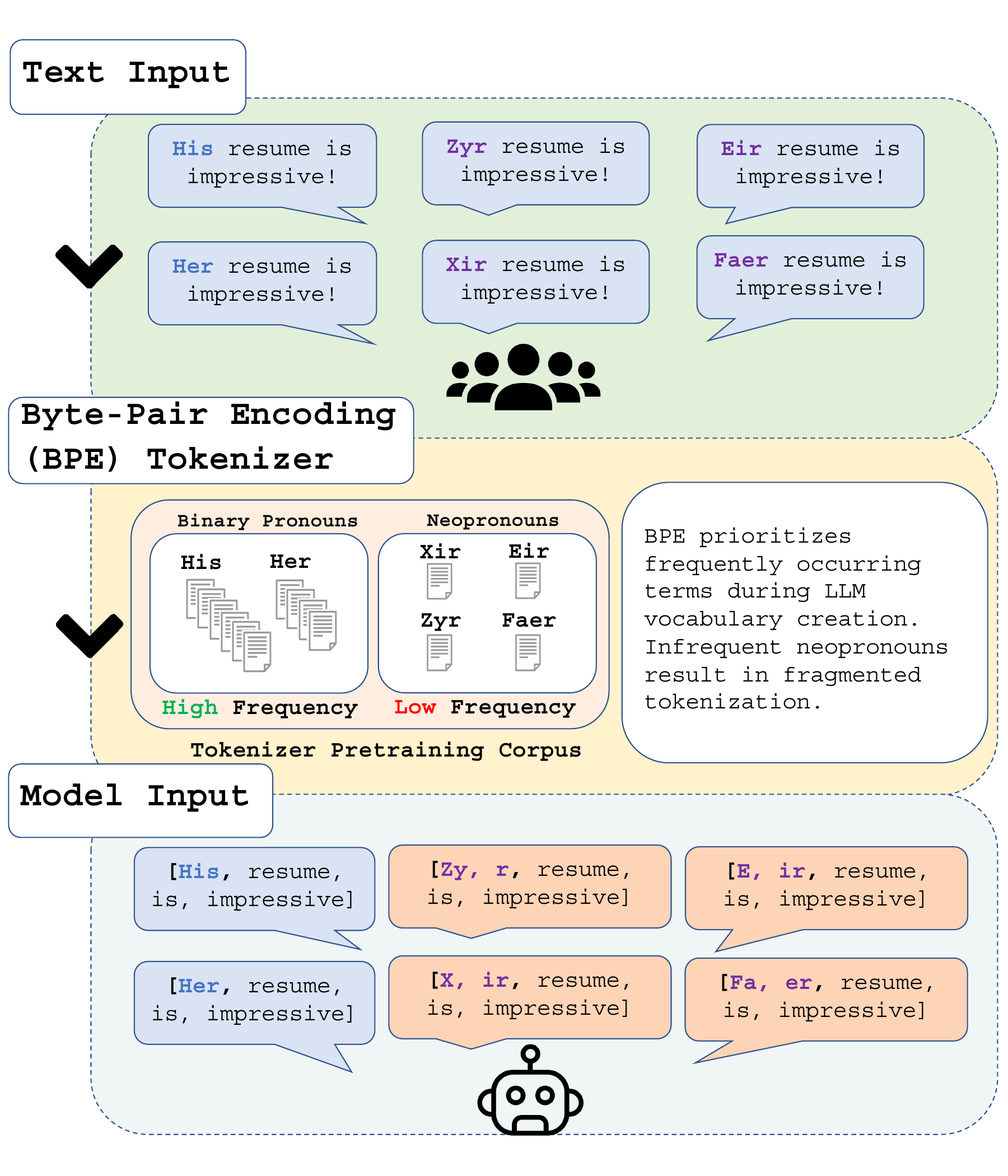}
    \vspace{-0.5cm}
    \caption{Byte-Pair Encoding (BPE) tokenization disproportionately fragments neopronouns compared to binary pronouns due to their infrequency in the training corpus. Our paper reveals that this overfragmentation leads to syntactic difficulties for LLMs, which are tied to their propensity to misgender data-scarce pronouns.}
    \label{fig:intro_teaser}
    \vspace{-0.1cm}
\end{figure}

\section{Introduction}


Gender bias in NLP has been extensively studied for binary gender, however mitigating harmful biases for underrepresented gender minorities remains an active area of research~\cite{sun2019mitigating, stanczak2021survey}.
Previous studies~\cite{dev-etal-2021-harms, ovalle2023m, hossain2023misgendered} have shown that large language models (LLMs) often fail to correctly use non-binary pronouns, particularly neopronouns such as \emph{xe} and \emph{ey}.
~\cite{sun2019mitigating, stanczak2021survey}.
Previous studies~\cite{dev-etal-2021-harms, ovalle2023m, hossain2023misgendered} have shown that large language models (LLMs) often fail to correctly use non-binary pronouns, particularly neopronouns such as \emph{xe} and \emph{ey}.\footnote{\url{https://nonbinary.wiki/wiki/English_neutral_pronouns}}
These works highlight the connection between LLM misgendering\footnote{The act of intentionally or unintentionally addressing someone (oneself or others) using a gendered term that does not match their gender identity.} and data scarcity, as neopronouns are severely underrepresented in pretraining corpora, thus limiting  the LLM's ability to use them proficiently. Despite this, the specific pathways through which data scarcity contributes to LLM misgendering behavior remain underexplored. Our work aims to address this research gap by investigating a critical, yet understudied aspect to LLM misgendering: tokenization.

\autoref{fig:intro_teaser} illustrates the tokenization differences between binary pronouns and neopronouns when using Byte-Pair Encoding (BPE), the most widely adopted subword tokenizer employed by popular LLMs such as GPT-4~\cite{brown2020language}, Claude \footnote{\url{https://www.anthropic.com/news/claude-3-family}}, Mistral~\cite{jiang2023mistral}, and Llama 2~\cite{touvron2023llama}. 
While binary pronouns (\textit{her} and \textit{his}) are tokenized as single units, neopronouns \textit{zyr}, \textit{eir}, \textit{xir}, and \textit{faer} are fragmented into two subword tokens due to their infrequency within the tokenizer's training corpus.
As a result, the LLM must rely on more granular subword tokens to learn the neopronoun's representation. Prior research finds that token overfragmentation adversely affects Part-of-Speech tagging and dependency parsing performance, as subword tokens share their embeddings across common words, introducing contextual ambiguity \citep{Wang2019ImprovingPM, limisiewicz2023tokenization}. However, the impact of this phenomenon on English LLM misgendering remains unexplored.

\paragraph{Contributions} To the best of our knowledge, our work is the first to link LLM misgendering to subword tokenization and deficient neopronoun grammar. We employ a series of evaluations that target understanding the association between LLM misgendering and poor pronoun morphosyntax (\S \ref{sec:metrics}), finding that neopronoun misgendering is strongly associated with an LLM's inability to use neopronouns as pronouns (\S \ref{sec:exp_measure}).

Through a series of carefully controlled experiments, we demonstrate that mitigations centered on improving LLM neopronoun proficiency reduce neopronoun misgendering. We introduce \textit{pronoun tokenization parity} (PTP), a technique to better preserve neopronoun tokens as functional morphemes by enforcing parity between neopronoun and binary pronoun tokenization (\S \ref{sec:PTP}). Furthermore, we investigate leveraging pre-existing LLM pronoun knowledge to improve the model's grammatical usage of neopronouns (\S \ref{sec:xlingual}). Our results demonstrate that finetuning GPT-based models with PTP achieves up to 58.4\% pronoun consistency, significantly outperforming the 14.1\% obtained from finetuning with standard BPE tokenization. Notably, finetuning the LLM's lexical layer with PTP outperforms traditional finetuning in 75\% of models, reducing compute time by up to 21.5\%. We find lexical finetuning consistently improves LLM pronoun consistency across model sizes, with smaller models experiencing the most significant gains—even matching the performance of models twice their size (\S\ref{sec:ablations}).

\section{Background}

\paragraph{Gender-Inclusive NLP} 
\indent
Gender bias has been studied across several NLP contexts, including machine translation~\cite{stanovsky2019evaluating}, coreference resolution~\cite{rudinger2018gender, Zhao2018GenderBI}, and named entity recognition~\cite{Mehrabi2019ManIT}. Works like \citep{gaido2021split} and others have found that choice of word segmentation exacerbates gender biases in machine translation.
Recent works expand gender bias evaluations to harms unique to non-normative gender communities within LLMs~\cite{dev-etal-2021-harms, hossain2023misgendered, ovalle2023m, nozza2022measuring, Felkner2023WinoQueerAC, QueerInAI2023BoundBT}. \citet{dev-etal-2021-harms} examine non-binary gender bias in static and contextual language representations, highlighting how data limitations affect these embeddings. Similarly, \citet{ovalle2023m} explore misgendering and harmful responses related to gender disclosure using their \texttt{TANGO} framework, pointing to challenges in neopronoun consistency, possibly due to data scarcity. \citet{hossain2023misgendered} corroborate these findings with an in-context-learning evaluation and analyses into LLM pretraining corpus statistics. 
Despite exploring various in-context learning strategies, they find persistent gaps between binary pronoun and neopronoun misgendering. These studies collectively emphasize data scarcity's impact on neopronouns, though questions remain regarding how data scarcity shapes neopronoun representations and subsequent LLM pronoun consistency.
In this study, we investigate the pivotal role of BPE tokenization due to its critical relationships to pretraining corpora and subsequent LLM vocabulary construction.

\paragraph{BPE Tokenization} 
Byte-Pair Encoding~\cite[BPE;][]{sennrich-etal-2016-improving} is a subword tokenization technique that constructs token vocabularies by iteratively merging frequently occurring adjacent token pairs up to a predefined vocabulary size.
Unseen or rare words are decomposed into subword units, down to individual characters, thus removing the need for assigning ``unknown'' token  (\texttt{[UNK]}) to unseen words. However, this approach does not consider context, posing limitations for task-relevant yet data-scarce scenarios~\citep{Yehezkel2022IncorporatingCI}.

\section{Low-Resource Challenges for BPE}
\label{sec:OOV}

\begin{table}
\footnotesize
\renewcommand{\arraystretch}{1.2} 
\setlength{\tabcolsep}{0.6pt} 
\centering
\begin{tabular}{@{}lcccccc@{}} 
\toprule
\multicolumn{1}{c}{\textbf{}}                                                            \textbf{}   & $\zeta$   & \textbf{Nom.}     & \textbf{Acc.}     & \begin{tabular}[c]{@{}c@{}}\textbf{Genitive}\\\textbf{ Dep.}\end{tabular} & \begin{tabular}[c]{@{}c@{}}\textbf{Genitive}\\\textbf{ Ind.}\end{tabular} & \textbf{Reflex.}          \\ 
\midrule
\multirow{2}{*}{\textbf{Binary}} & \multirow{2}{*}{1.20} & he                & him               & his                                                                       & his                                                                       & {[}him, self]             \\
                                                                                            &                       & she               & her               & her                                                                       & hers                                                                      & {[}her, self]             \\ 
\midrule
\multirow{9}{*}{\begin{tabular}[c]{@{}l@{}}\textbf{Neo}\\\textbf{}\end{tabular}}   & \multirow{9}{*}{1.87} & ey                & em                & \textbf{[ei, r]}                                                          & \textbf{[e, irs]}                                                         & {[}em, self]              \\
                                                                                            &                       & xe                & \textbf{[x, em]}  & \textbf{[x, ir]}                                                          & \textbf{[x, irs]}                                                         & \textbf{[x, ir, self]}    \\
                                                                                            &                       & \textbf{[f, ae]}  & \textbf{[fa, er]} & \textbf{[fa, er]}                                                         & \textbf{[fa, ers]}                                                        & \textbf{[fa, ers, elf]}   \\
                                                                                            &                       & zie               & \textbf{[z, ir]}  & \textbf{[z, ir]}                                                          & \textbf{[z, irs]}                                                         & \textbf{[z, ir, self]}    \\
                                                                                            &                       & ze                & \textbf{[h, ir]}  & \textbf{[h, ir]}                                                          & \textbf{[h, irs]}                                                         & \textbf{[h, ir, self]}    \\
                                                                                            &                       & sie               & \textbf{[h, ir]}  & \textbf{[h, ir]}                                                          & \textbf{[h, irs]}                                                         & \textbf{[h, ir, self]}    \\
                                                                                            &                       & \textbf{[th, on]} & \textbf{[th, on]} & \textbf{[th, ons]}                                                        & \textbf{[th, ons]}                                                        & \textbf{[th, ons, self]}  \\
                                                                                            &                       & ve                & ver               & vis                                                                       & vis                                                                       & {[}vers, elf]             \\
                                                                                            &                       & ne                & ner               & \textbf{[n, is]}                                                          & \textbf{[n, is]}                                                          & {[}nem, self]             \\
\bottomrule
\end{tabular}
\caption{BPE-tokenized Binary Pronouns and Neopronouns across pronoun forms. $\zeta$= Fertility. The closer fertility is to 1, the more the tokenizer kept pronoun tokens fully intact. \textbf{Bold} = neopronoun tokenization that does not follow binary pronoun forms.}
\label{tbl:token}
\end{table}
\paragraph{Data-Scarce Tokenization}

\label{sec:consequences_OOV}

\citet{bostrom2020byte} find that tokenization introduces a significant amount of inductive bias in LLMs, profoundly impacting their ability to perform tasks downstream.
BPE prioritizes keeping the most frequent words intact during tokenization while splitting lower-frequency texts into smaller subword tokens, irrespective of their contextual relevance~\citep{Yehezkel2022IncorporatingCI, mielke2021between}.
This behavior leads to learning critical aspects of language, like pronoun morphosyntax, through reliance on textual frequency, resulting in a fragmented understanding of morphosyntactic rules for less frequent pronoun sets. This tokenization disparity is reflected in \autoref{tbl:token} across tokenized pronoun groups and their respective fertility scores~\citep{rust-etal-2021-good}, i.e., the average number of subwords produced per tokenized word. Binary pronouns are kept intact after tokenization, while most neopronouns are segmented into subword tokens, indicating that the LLM's predefined vocabulary cannot construct these tokens. We posit that this lack of parity in tokenization between pronouns contributes to LLM misgendering downstream.

\paragraph{OOV Pronouns and Hindered Grammatical Knowledge}

\citet{Wang2019ImprovingPM} find that \textit{OOV words}, words that were unable to remain fully intact after tokenization, have detrimental impacts on downstream part-of-speech (POS) proficiency.
Resulting token overfragmentation presents challenges across additional tasks such as named entity recognition~\cite{Daena2020QualityOW, Wang2022MINERIO}, dependency parsing ~\cite{limisiewicz2023tokenization}, and machine translation~\cite{Domingo2018HowMD, Huck2019BetterOT, Araabi2022HowEI}.
\citet{limisiewicz2023tokenization} find that because subwords are present in multiple words, their embeddings incorporate information from these common words, making the resulting ambiguity challenging to parse. Because of this, we hypothesize that the observed overfragmentation of tokenized neopronouns relates to LLM deficiencies in learning proper neopronoun morphosyntax.

\section{Tracing LLM Misgendering to Grammatical Deficiencies}
\label{sec:metrics}

\begin{figure}[t]
    \includegraphics[width=1\linewidth]{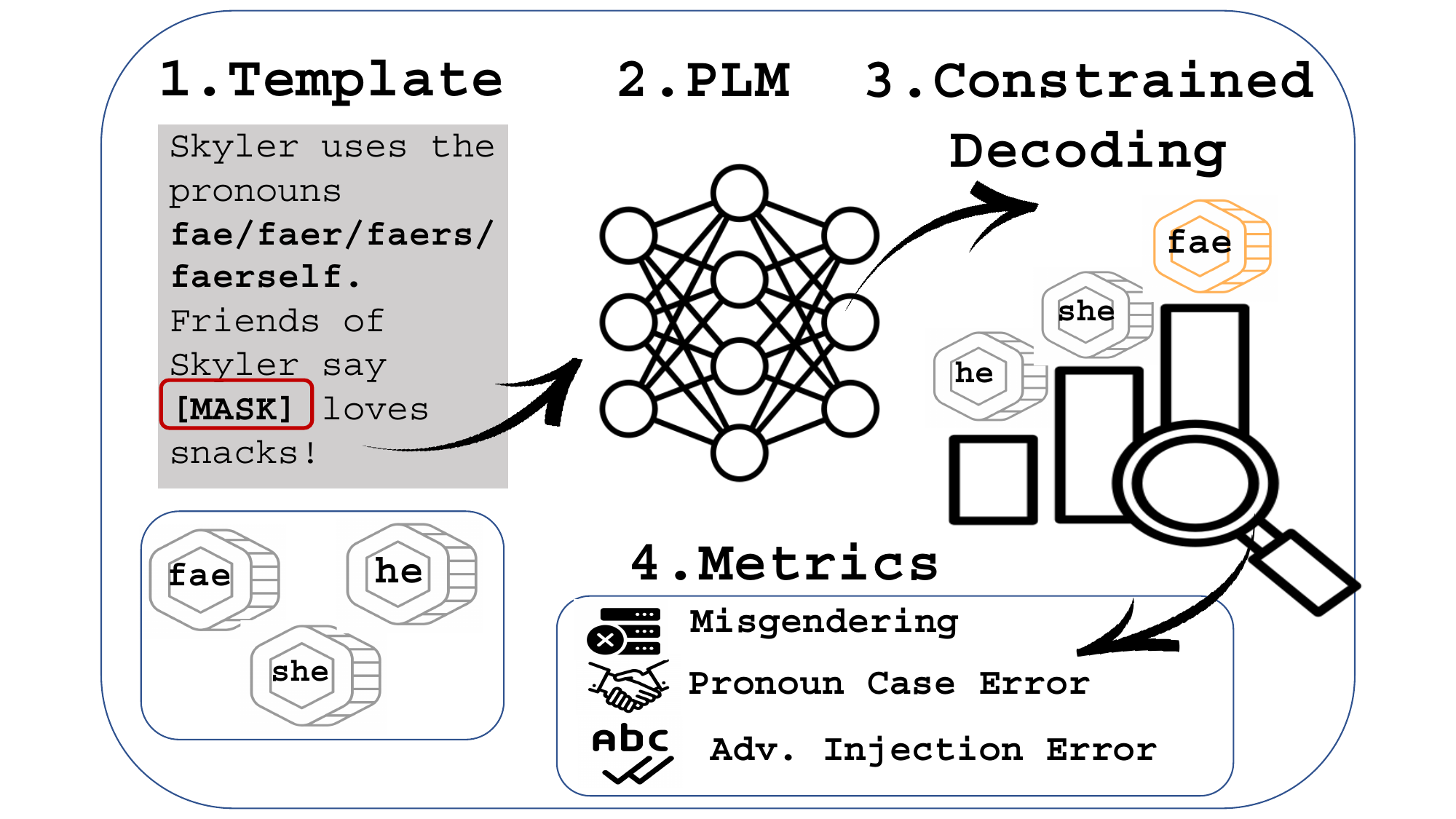}
    \vspace{-0.5cm}
    \caption{Evaluation. We determine our method's efficacy in reducing LLM misgendering using a constrained decoding approach across 3 metrics.}   
\label{fig:eval}
\end{figure}

This section presents a series of metrics to evaluate LLM misgendering from the standpoint of pronoun proficiency. We perform baseline evaluations on out-of-the-box GPT-Neo-X based models and provide an overview of our evaluation scheme in \autoref{fig:eval}.

\subsection{Evaluation Setup}
\label{sec:metrics_eva_setup}

\paragraph{Models}
We employ the Pythia model suite for our evaluation and experiments,\footnote{\url{https://github.com/EleutherAI/pythia}} as it parallels state-of-the-art architecture; Pythia models are all built on top of a GPT-Neo-X architecture, an open-source alternative to GPT-3 models. Notably, it is based on a BPE tokenizer~\cite{Biderman2023PythiaAS} and trained on the \textsc{PILE} dataset \citep{gao2020pile}.

\paragraph{Dataset}
We utilize the \texttt{MISGENDERED} dataset by \citet{hossain2023misgendered}, containing added templates and names from \texttt{TANGO} \cite{ovalle2023m}, resulting in 93,600 templates to evaluate LLMs on our three metrics. We provide further dataset details in the sections below and in the Appendix (\S\ref{app:more_templates}).

\subsection{Evaluation Metrics}
\label{sec:metrics_eval}

According to \citet{garner2016chicago}, English pronouns must agree with their subject in gender, case, and number. We define three metrics to quantify a model's understanding of different pronoun forms: two are standard misgendering measurements, and one is a novel metric introduced in this paper. 
\textit{Pronoun consistency} (Consistency) assesses pronoun-gender agreement and is the primary metric for determining performance improvement in this paper. 
Previous studies find that this automatic consistency evaluation highly correlates to human evaluation \citep{ovalle2023m}. \textit{Pronoun Case Agreement Error} (Case Error) is an auxiliary metric that provides insight into how well the model has learned pronoun forms. To test the relationship between LLM misgendering and poor LLM morphosyntax, we introduce \textit{Adversarial Injection Error} (Inject Error) to measure LLM robustness against word insertion adversarial attacks that render a sentence grammatically incorrect or change its meaning. If there is an association between poor consistency and adversarial error, it would support formulating mitigations that prioritize enhancing the LLM's overall grammatical proficiency with neopronouns.
These metrics are employed in a constrained decoding setting, consistent with the \texttt{MISGENDERED} framework introduced by \citet{hossain2023misgendered}. Given a masked template, the LLM predicts the most likely pronoun from a pool of pronouns of the same form.

\paragraph{Pronoun Consistency}
Let $S$ be a set of unique pronoun families with $|S|$ pronoun families. Each pronoun family $M$ $\in$ $S$ contains $|M|$ English pronoun forms.
Within a collection of masked templates $T$, \texttt{[MASK]} is replaced with a pronoun $p \in M$ for all $M \in S$, resulting in the filled template set $T^*$. In line with \citet{hossain2023misgendered}, each template starts with a person's name and their pronoun declaration (i.e., nominative / accusative / genitive / reflexive), followed by a sentence containing a \texttt{[MASK]} token which expects a pronoun.
For example: \textit{Casey uses the pronouns he/him/his/himself. Upon recognizing Casey, the fan asked \texttt{[MASK]} for an autograph.}. For a template $t$ consisting of $m$ tokens $x_1, x_2, \ldots, x_m$, the token generated at \texttt{[MASK]}, $\hat{y}_{t}$, is defined as the  \texttt{argmax} transition probability from the pronoun pool. 

\vspace{-0.25cm}
\begin{equation}
\hat{y}_{t} = \texttt{argmax}_{p \in S} P(x_{i} = s | x_{<i})
\label{eq:argmax}
\end{equation}

We denote the set of filled templates as $C$. Each filled template is then compared to its golden label example $c \in C^*$, containing the correct pronoun for that template-name-declaration combination.  

To evaluate pronoun consistency, we compare the model's chosen pronoun for a template, $\hat{y}_{t}$, to the template's correct pronoun, $y_{c}$, and then calculate the accuracy over all templates: 

\begin{equation}
\frac{1}{|T^*|} \sum_{t \in T^*, y \in C^*} \delta(\hat{y}_{t}, y_{c})
\label{eq:compare}
\end{equation}

\paragraph{Pronoun Case Error}
Evaluating pronoun case error is essential for assessing a model's competence in pronoun usage.
Ideally, an LLM would generate case-agreeing sentences like ``She went to the store.'' instead of ``Hers went to the store.''
To evaluate this, we use the same approach as above, instead focusing on assessing expected versus predicted pronoun cases for a given pronoun family. 
However, transition probabilities conditioned solely on preceding tokens cannot be relied on to determine case correctness.
For example, a sentence like ``Casey went to the store for \texttt{[MASK]} mom'' can have its mask replaced with ``her'' or ``herself'' and still be grammatically correct, as it only considers the previous tokens during inference.
Therefore, we obtain the model's predicted output across all pronoun cases for a given family $s \in Q$, minimizing its loss (i.e., maximizing probability). Pronoun case error is then the proportion of templates with \textit{incorrect} case agreement for a given pronoun family.

\vspace{-0.3cm}
\begin{equation}
\texttt{argmin}_{s \in Q} \left(-\sum_{i=1}^{N}\log P_{\theta}(x_{i}| x_{<i})\right)
\label{eq:argmin}
\end{equation}

\paragraph{Adversarial Injection Error}
Prior research finds that prompting LLMs with texts containing neopronouns often results in ungrammatical generations, where neopronouns are incorrectly preceded by articles and determiners such as 'the', 'a', or 'these'~\citep{ovalle2023m}.  To further examine an LLM's inability to construct grammatically correct sentences with neopronouns, we replicate this observed behavior by generating a set of otherwise grammatically correct prompts that include adversarial word insertions, making the template entirely ungrammatical.
We use the same templates as previously defined but now augment each \texttt{[MASK]} to \texttt{[DET]\textvisiblespace [MASK]}, where \texttt{[DET]} is replaced by singular and plural determiners (e.g., `this', `those', `these'), articles (like `the', `a'), or no determiner at all. Example templates are provided in Appendix~\ref{app:more_templates}. Similar to pronoun consistency, we employ LLM transition probabilities to evaluate how often LLMs use neopronouns in ungrammatical contexts.
Next, we analyze the LLM's output by calculating the \texttt{argmax} of the transition probability for all potential substitutions of \texttt{[DET]} (\autoref{eq:argmax}).
An LLM utilizing a neopronoun correctly should choose a template without a determiner. Models displaying incorrect behavior indicates poor grammatical proficiency with neopronouns.

\subsection{Results}
\label{sec:exp_measure}

We report pronoun consistency, pronoun case error, and adversarial injection errors in \autoref{tbl:intro_metrics}. In line with prior work, the neopronoun \textit{xe} reflects the lowest pronoun consistency (i.e., highest misgendering) across all model sizes. To better understand how this relates to grammatical issues, we also calculate Spearman's correlation between pronoun consistency and each of the two error metrics (leftmost results column). Notably, we observe moderate to strong negative correlations between grammatical error metrics and misgendering, with adversarial injections most strongly correlated. Across model sizes, we find a range of $-$0.45 to $-$0.63 correlation for injection error and $-$0.53 to $-$0.63 for case error. With these observations, we posit that mitigation strategies that enhance an LLM's grammatical proficiency with neopronouns will attenuate their tendency to misgender.

\begin{table}[!t]
\centering
\footnotesize
\renewcommand{\arraystretch}{1.2} 
\setlength{\tabcolsep}{2.1pt} 
\begin{tabular}{clcccc} 
\toprule
\multirow{2}{*}{\textbf{Size}} & \multicolumn{1}{c}{\multirow{2}{*}{\textbf{Metric}}} & \multirow{2}{*}{\textbf{$\rho$}} & \multicolumn{3}{c}{\textbf{Pronoun Family}}                                                 \\
                               & \multicolumn{1}{c}{}                                 &                                  & \textbf{He}               & \textbf{She}              & \textbf{Xe}                         \\ 
\midrule
\multirow{3}{*}{70M}           & \textbf{Consistency ($\uparrow$)}                    & —                                & 96.82\textsubscript{0.77} & 71.59\textsubscript{2.00} & \textbf{0.67\textsubscript{0.35}}   \\
                               & \textbf{Case Error ($\downarrow$)}                   & -0.63                            & 8.26\textsubscript{1.21}  & 24.36\textsubscript{1.90} & \textbf{78.56\textsubscript{1.82}}  \\
                               & \textbf{Inject Error ($\downarrow$)}                 & -0.45                            & 23.85\textsubscript{1.88} & 16.92\textsubscript{1.66} & \textbf{85.03\textsubscript{1.58}}  \\ 
\midrule
\multirow{3}{*}{160M}          & \textbf{Consistency ($\uparrow$)}                    & —                                & 79.95\textsubscript{1.82} & 76.46\textsubscript{1.90} & \textbf{0.00\textsubscript{0.00}}   \\
                               & \textbf{Case Error ($\downarrow$)}                   & -0.59                            & 4.05\textsubscript{0.90}  & 10.87\textsubscript{1.38} & \textbf{80.00\textsubscript{1.77}}  \\
                               & \textbf{Inject Error ($\downarrow$)}                 & -0.63                            & 8.72\textsubscript{1.28}  & 6.46\textsubscript{1.10}  & \textbf{95.38\textsubscript{0.92}}  \\ 
\midrule
\multirow{3}{*}{410M}          & \textbf{Consistency ($\uparrow$)}                    & —                                & 72.82\textsubscript{1.92} & 55.85\textsubscript{2.21} & \textbf{0.05\textsubscript{0.08}}   \\
                               & \textbf{Case Error ($\downarrow$)}                   & -0.53                            & 2.87\textsubscript{0.74}  & 7.90\textsubscript{1.21}  & \textbf{79.90\textsubscript{1.79}}  \\
                               & \textbf{Inject Error ($\downarrow$)}                 & -0.54                            & 4.15\textsubscript{0.90}  & 3.49\textsubscript{0.79}  & \textbf{89.85\textsubscript{1.36}}  \\ 
\midrule
\multirow{3}{*}{1.4B}          & \textbf{Consistency ($\uparrow$)}                    & —                                & 78.46\textsubscript{1.82} & 66.56\textsubscript{2.03} & \textbf{0.26\textsubscript{0.23}}   \\
                               & \textbf{Case Error ($\downarrow$)}                   & -0.54                            & 3.54\textsubscript{0.82}  & 3.03\textsubscript{0.74}  & \textbf{76.00\textsubscript{1.92}}  \\
                               & \textbf{Inject Error ($\downarrow$)}                 & -0.62                            & 3.69\textsubscript{0.85}  & 3.44\textsubscript{0.79}  & \textbf{92.77\textsubscript{1.15}}  \\
\bottomrule
\end{tabular}
\caption{Out-of-the-box evaluations on Pythia, a GPTNeo-X based model across sizes. Uncertainty estimates are 95\% confidence intervals computed from 10k bootstrap iterations. \textit{Takeaway: Markedly higher grammatical error rates for neopronoun vs. binary pronouns.}}
\label{tbl:intro_metrics}
\end{table}

\section{Improving LLM Neopronoun Proficiency}
\label{sec:approach}

\begin{figure*}[!t]
    \centering
    \includegraphics[width=\textwidth]{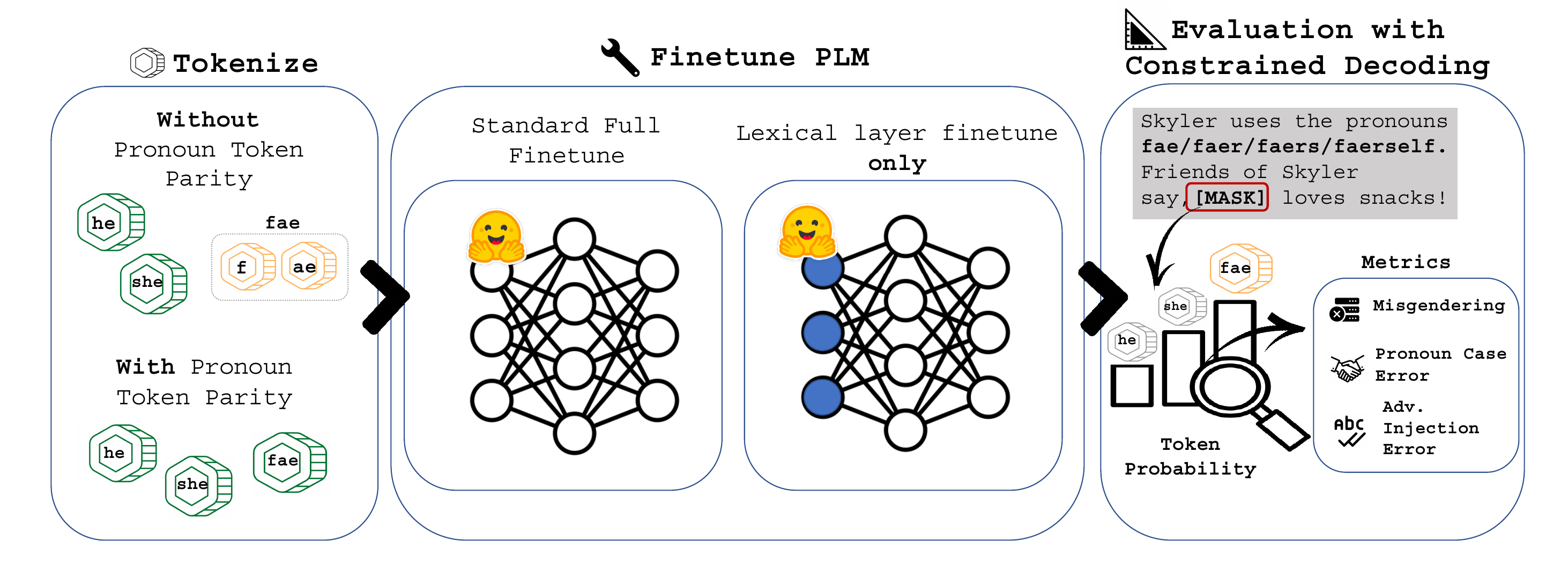}
    \vspace{-0.2cm}
    \caption{Overview. We (1) tokenize neopronouns using PTP for a given LLM, (2) either fully finetune or only finetune the LLM lexical layer with data containing neopronouns, and (3) determine our method's efficacy in reducing LLM misgendering using a constrained decoding approach across 3 metrics. }  
    \vspace{-0.1cm}  
\label{fig:motivation}
    
\end{figure*}

\subsection{Pronoun Tokenization Parity}
\label{sec:PTP}

English pronouns serve as building blocks for language acquisition. Termed \textit{functional morphemes}, these small, self-contained units of meaning reflect specific English grammatical functions~\cite{fortescue2005historical,morph2-slides}. To improve LLM neopronoun consistency, we introduce \textit{pronoun tokenization parity} (PTP), a method that maintains a token's functional integrity during BPE tokenization. By aligning neopronoun tokenization with that of binary pronouns, we aim to improve an LLM's grammatical understanding of neopronouns, ultimately enhancing the model's ability to use them correctly. 

Formally, we extend the pretrained token embeddings of a transformer-based LLM $E_{1}^{\text{orig}}, E_{2}^{\text{orig}}, \ldots, E_{n}^{\text{orig}}$, where $n$ represents the vocabulary size of the original model.
We introduce new embeddings $E^{\text{PTP}}$ for each of $m$ unique pronouns in the set of neopronoun cases (i.e., pronoun family) $S$, resulting in an extended vocabulary: $\{E_{1}^{\text{orig}}, \ldots, E_{n}^{\text{orig}}\} \cup \{ E_{1}^{\text{PTP}}, \ldots, E_{m}^{\text{PTP}}\}$. We provide additional details and instructions for reproducing PTP in Algorithm~\ref{app:algo}.

\subsection{Leveraging LLM Pre-Existing Pronoun Knowledge}
\label{sec:xlingual}

Training a new tokenizer and LLM requires significant computational resources and data. Pre-trained English LLMs have learned English syntax and pronouns during pretraining.
We can take advantage of morphosyntactic similarities between binary pronouns and neopronouns, such as their syntactic roles and agreement patterns, to transfer knowledge from one set of pronouns to another. 

Guided by fundamental aspects of cross-lingual transfer detailed in 
\citet{Artetxe2019OnTC} and ~\citet{Vries2020AsGA}, we propose the practice of finetuning only an LLM's lexical embedding layer while keeping downstream transformer weights fixed. 
As long as the source and target pronoun groups share similar linguistic foundations, mirroring those found in cross-lingual sharing of basic elements, we can sidestep common challenges in cross-lingual transfer, such as determining the most suitable transfer source language.
Unlike \citet{Artetxe2019OnTC}, we forgo training the transformer weights after freezing lexical embeddings since the new tokens already align with English grammar and syntax, eliminating the need for the transformer to adapt to a different language. Furthermore, in contrast to the approach by \citet{Vries2020AsGA}, we avoid resetting the entire lexical embedding layer to preserve the prelearned English grammar dependencies.

\section{Experimental Setup}

We provide an overview of our experimental setup in \autoref{fig:motivation}. We conduct carefully controlled experiments across two finetuning paradigms using open-source LLMs that vary in model size and neopronoun data scarcity. In the first set of experiments, we employ PTP in a standard full finetuning paradigm. In the second experiment, we introduce lexical finetuning and variants with  PTP.
We perform these experiments across binary pronouns and the neopronoun family \textit{xe}. We center \textit{xe} for several reasons: \textit{xe} ranks among the most widely adopted non-binary pronouns \cite{gendercensus2023}. 
Non-binary pronouns also exhibit diverse linguistic variations, spanning from closed to open word class forms \cite{miltersen2016nounself, lauscher2022welcome}.
This diversity requires a nuanced yet flexible approach.
By focusing on the \textit{xe} pronoun family, we showcase the effectiveness of PTP while providing a generalizable framework for researchers to build upon for studying non-binary pronouns within their respective linguistic contexts.

\subsection{Finetuning Dataset}
We finetune our models on the \textsc{Wikibios}\footnote{\url{https://huggingface.co/datasets/wiki_bio}} dataset, comprising 728,321 English biographical texts from Wikipedia.
Counterfactual data augmentation is used to address the limited availability and narrow dimensions of textual corpora containing neopronouns. 
We replace a variable proportion of binary pronouns with their neopronoun counterparts. Acknowledging that individuals who use neopronouns often have prior associations with binary pronouns, this data curation strategy enables LLMs to acquire knowledge of neopronouns within more comprehensive, diverse, and real-world contexts~\cite{Talat2022BackTT}.

We filter the \textsc{Wikibios} dataset to retain texts containing binary pronouns, resulting in 462,345 examples.
Each binary pronoun is replaced with its corresponding neopronoun case, incorporating correct possessive forms using the spaCy part-of-speech tagger.\footnote{\url{https://spacy.io/}}
No biography text appears more than once in the dataset splits.

To understand how our methods operate across data resource levels, we counterfactually augment with an increasing proportion of neopronouns: 10\%, 20\%, 30\%, 40\%, and 50\%. At the 50\% level, the dataset is evenly split between neopronouns and binary pronouns.

\subsection{Finetuning Setups}

\paragraph{Pronoun Tokenization Parity}
To test whether PTP helps mitigate LLM misgendering, we prepare two versions of finetuning for a compact 70M parameter Pythia model. The first model is finetuned with its original BPE tokenizer (\textsc{$\text{T}_{\text{Orig}}$}) and the second with PTP (\textsc{$\text{T}_{\text{PTP}}$}).
Embeddings for \textsc{$\text{T}_{\text{PTP}}$} are initialized with a random Gaussian ($\mu$=0 and $\sigma$=0.02).
\textsc{$\text{M}_{\text{Full}}$} denotes all models with standard full finetuning, and \textsc{$\text{M}_{\text{Base}}$} represents the HuggingFace out-of-the-box checkpoint which uses its original BPE tokenizer \textsc{$\text{T}_{\text{Orig}}$}. \textsc{$\text{T}_{\text{Orig}}$}+ \textsc{$\text{M}_{\text{Base}}$} and \textsc{$\text{T}_{\text{Orig}}$}+ \textsc{$\text{M}_{\text{Full}}$} serve as baselines for PTP.

Each model is finetuned across five epochs with a batch size of 128 and a $10^{-4}$ learning rate. We employ several techniques to encourage model generalization and prevent overfitting. We incorporate weight decay regularization (0.01), a warmup ratio of 0.01 to gradually increase the learning rate over the initial 1\% of training steps, and apply early stopping based on cross-entropy loss in the validation set with a patience of 2. All models undergo finetuning using FP16 mixed precision and two gradient accumulation steps. We provide further details on our setup in Appendix~\ref{app:model_details}.

\paragraph{Lexical Layer Finetuning}
We follow the same setup as before but now increase the learning rate to $10^{-3}$ to encourage more rapid adaptation to the new vocabulary. We denote models trained with lexical finetuning with original BPE tokenization as \textsc{$\text{T}_{\text{Orig}}$}+ \textsc{\textsc{$\text{M}_{\text{Lex}}$}}.
We compare performance to PTP and PTP baselines:  \textsc{$\text{T}_{\text{PTP}}$}+ \textsc{$\text{M}_{\text{Full}}$}, \textsc{$\text{T}_{\text{Orig}}$}+ \textsc{$\text{M}_{\text{Base}}$} and \textsc{$\text{T}_{\text{Orig}}$}+ \textsc{$\text{M}_{\text{Full}}$}. We also introduce an additional lexical finetuning variant with PTP (\textsc{$\text{T}_{\text{PTP}}$}+ \textsc{\textsc{$\text{M}_{\text{Lex}}$}}) and test to what extent combining these techniques boosts performance over either method.

\paragraph{Model Size Ablations}
In order to evaluate the effectiveness of our proposed mitigations at various scales and resource levels, we repeat our experiments at 160M, 410M, and 1.4B parameters. Furthermore, we ensure that all finetuned models do not overfit nor adversely impact pre-existing performance on downstream tasks, reporting test set evaluations and a case study on downstream tasks in Appendix~\ref{sec:training_evalconf} and~\ref{app:evaluations}.




\section{Results}

\begin{table}[t]
\centering
\footnotesize
\renewcommand{\arraystretch}{1} 
\setlength{\tabcolsep}{2.5pt} 
\begin{tabular}{clccc} 
\toprule
\textbf{Model}                                                                                                                                                                                            & \multicolumn{1}{c}{\textbf{Metric}}                      & \textbf{He}                        & \textbf{She}                       & \textbf{Xe}                         \\ 
\midrule
\multirow{3}{*}{\begin{tabular}[c]
{@{}l@{}}\textbf{{$\text{T}_{\text{Orig}}+$}}\\\textbf{$\text{M}_{\text{Base}}$}\end{tabular}}& \textbf{Consistency ($\uparrow$)}    & \textbf{96.82\textsubscript{0.79}} & 71.59\textsubscript{2.03}          & 0.67\textsubscript{0.38}            \\
                                                                                                                                                                                                          & \textbf{Case Error ($\downarrow$)}   & 8.26\textsubscript{1.26}           & 24.36\textsubscript{1.90}          & 78.56\textsubscript{1.77}           \\
                                                                                                                                                                                                          & \textbf{Inject Error ($\downarrow$)} & 23.85\textsubscript{1.90}          & 16.92\textsubscript{1.67}          & 85.03\textsubscript{1.56}           \\ 
\midrule
\multirow{3}{*}{\begin{tabular}[c]
{@{}l@{}}\textbf{{$\text{T}_{\text{Orig}}+$}}\\\textbf{$\text{M}_{\text{Full}}$}\end{tabular}}        & \textbf{Consistency ($\uparrow$)}    & 89.64\textsubscript{1.36}          & \textbf{86.05\textsubscript{1.54}} & 14.46\textsubscript{1.56}           \\
                                                                                                                                                                                                          & \textbf{Case Error ($\downarrow$)}   & 11.74\textsubscript{1.44}          & 22.41\textsubscript{1.87}          & 59.95\textsubscript{2.15}           \\
                                                                                                                                                                                                          & \textbf{Inject Error ($\downarrow$)} & 23.95\textsubscript{1.87}          & 16.77\textsubscript{1.67}          & 89.49\textsubscript{1.36}           \\ 
\midrule
\multirow{3}{*}{\begin{tabular}[c]
{@{}l@{}}\textbf{{$\text{T}_{\text{PTP}}+$}}\\\textbf{$\text{M}_{\text{Full}}$}\end{tabular}}        & \textbf{Consistency ($\uparrow$)}    & 94.77\textsubscript{0.97}          & 83.49\textsubscript{1.67}          & 37.79\textsubscript{2.10}           \\
                                                                                                                                                                                                          & \textbf{Case Error ($\downarrow$)}   & 9.69\textsubscript{1.31}           & 29.28\textsubscript{2.00}          & 56.92\textsubscript{2.15}           \\
                                                                                                                                                                                                          & \textbf{Inject Error ($\downarrow$)} & 27.79\textsubscript{1.95}          & 20.97\textsubscript{1.79}          & 27.03\textsubscript{1.95}           \\ 
\midrule
\multirow{3}{*}{\begin{tabular}[c]
{@{}l@{}}\textbf{{$\text{T}_{\text{Orig}}+$}}\\\textbf{$\text{M}_{\text{Lex}}$}\end{tabular}}          & \textbf{Consistency ($\uparrow$)}    & 86.46\textsubscript{1.49}          & 72.87\textsubscript{2.00}          & 16.77\textsubscript{1.62}           \\
                                                                                                                                                                                                          & \textbf{Case Error ($\downarrow$)}   & 18.51\textsubscript{1.72}          & 33.79\textsubscript{2.08}          & 70.51\textsubscript{2.05}           \\
                                                                                                                                                                                                          & \textbf{Inject Error ($\downarrow$)} & 28.97\textsubscript{2.05}          & 23.18\textsubscript{1.87}          & 65.44\textsubscript{2.10}           \\ 
\midrule
\multirow{3}{*}{\begin{tabular}[c]
{@{}l@{}}\textbf{{$\text{T}_{\text{PTP}}+$}}\\\textbf{$\text{M}_{\text{Lex}}$}\end{tabular}}          & \textbf{Consistency ($\uparrow$)}    & 84.97\textsubscript{1.59}          & 72.21\textsubscript{1.95}          & \textbf{53.59\textsubscript{2.21}}  \\
                                                                                                                                                                                                          & \textbf{Case Error ($\downarrow$)}   & 18.15\textsubscript{1.72}          & 33.03\textsubscript{2.08}          & 60.46\textsubscript{2.15}           \\
                                                                                                                                                                                                          & \textbf{Inject Error ($\downarrow$)} & 25.79\textsubscript{1.97}          & 21.85\textsubscript{1.82}          & 34.77\textsubscript{2.10}           \\
\bottomrule
\end{tabular}

\caption{70M-parameter model results at 10\% data resource level.  \textsc{$\text{T}_{\text{Orig}}$}= original BPE tokenizer, \textsc{$\text{T}_{\text{PTP}}$}= tokenizer with \texttt{PTP},  \textsc{$\text{M}_{\text{Base}}$}= original model (no finetuning) \textsc{$\text{M}_{\text{Full}}$}= full finetuning. Uncertainty estimates are 95\% confidence intervals computed from 10k bootstrap iterations. }
\label{tbl:70m_model_comparison}
\end{table}

\begin{figure}[!t]
    \centering
    \includegraphics[width=\linewidth]{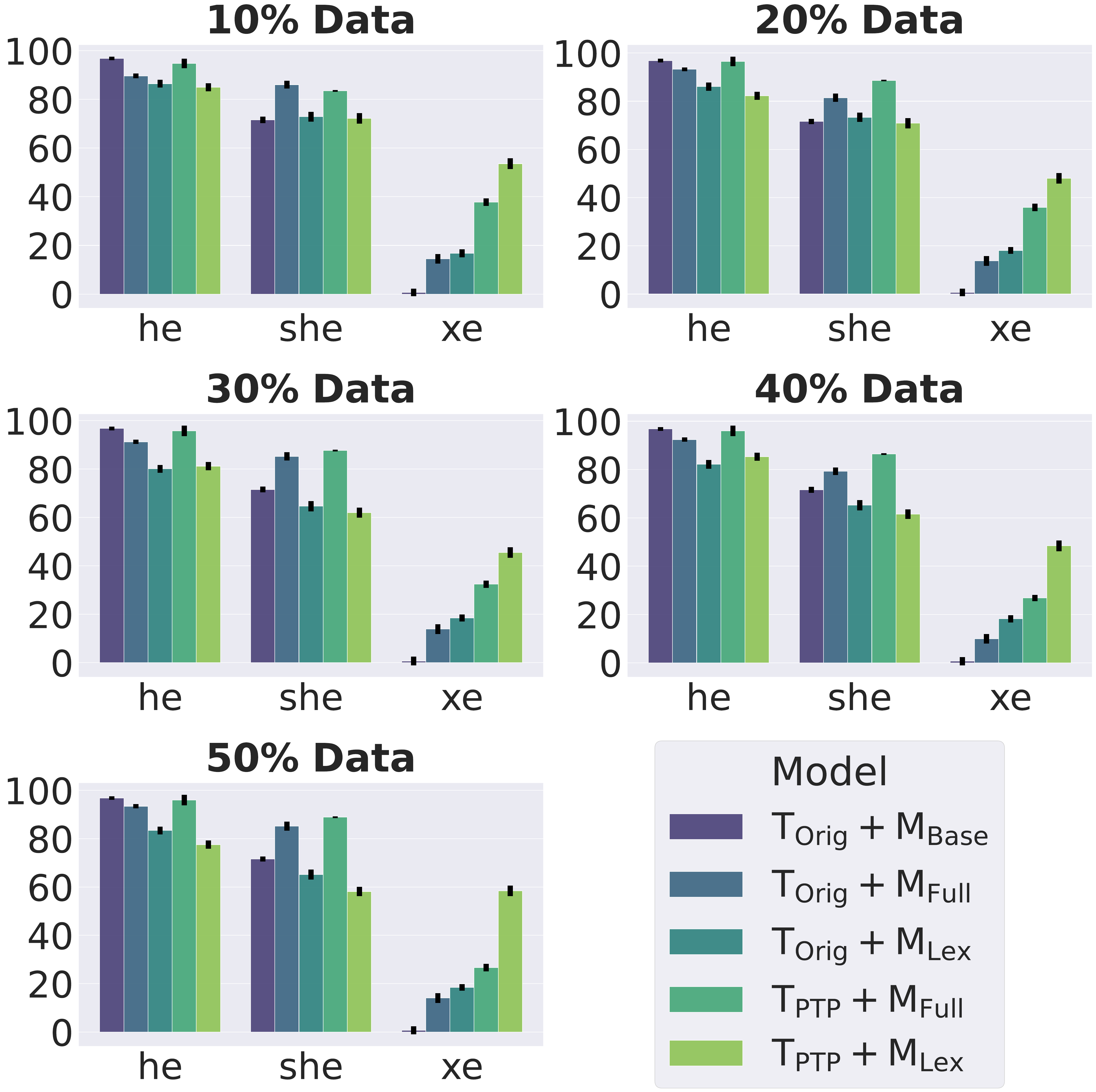}
    \caption{70M model pronoun consistency for each pronoun family  across 10-50\% data resource levels and model variants. \textit{Takeaway: PTP sustains improvements in neopronoun consistency across data resource levels.}}
    \label{fig:70m_hf}

\end{figure}

\begin{figure*}[!t]
    \centering
    \includegraphics[width=\linewidth]{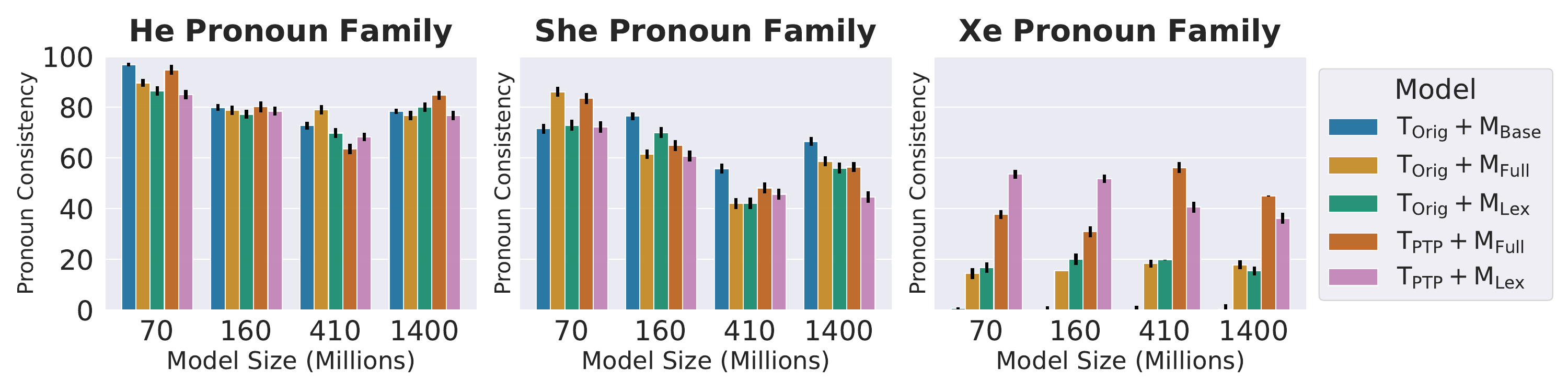}
    \caption{Results across all models at data resource level=10. The uncertainty estimates are 95\% confidence intervals computed from 10k bootstrap iterations. \textit{Takeaway: Across model size, variants of PTP consistently improve neopronoun consistency over models employed with standard BPE (Baseline, \textsc{$\text{T}_{\text{PTP}} + \text{M}_{\text{Full}}$}}).} 
    \vspace{-.5cm}
    \label{fig:all_model_size}
\end{figure*}

\subsection{Pronoun Tokenization Parity}

We report our PTP finetuning results in \autoref{tbl:70m_model_comparison}. Both \textsc{$\text{T}_{\text{PTP}} + \text{M}_{\text{Full}}$} (37.8\%) and \textsc{$\text{T}_{\text{Orig}} + \text{M}_{\text{Full}}$} (14.5\%) demonstrated gains in neopronoun consistency over \textsc{$\text{T}_{\text{Orig}} + \text{M}_{\text{Base}}$} (<1\%).
This improvement is expected, considering their increased exposure to neopronouns during finetuning. 
However, \textbf{models using PTP outperformed those finetuned with original BPE tokenization}.
As shown in \autoref{fig:70m_hf}, PTP's improvement over these two baselines was consistent across data resource levels. We observed the best neopronoun consistency overall at 58.4\% (50\% data resource level).
Notably, gains over vanilla finetuning (\textsc{$\text{T}_{\text{Orig}} + \text{M}_{\text{Full}}$}) were most evident at resource levels below 30\%, where \textsc{$\text{T}_{\text{PTP}} + \text{M}_{\text{Full}}$} more than doubled neopronoun consistency over \textsc{$\text{T}_{\text{Orig}} + \text{M}_{\text{Full}}$} (14.5\% vs. 37.8\%).
Binary pronoun consistency remained stable, with \textsc{$\text{T}_{\text{PTP}} + \text{M}_{\text{Full}}$} even improving \textit{she} pronoun consistency over \textsc{$\text{T}_{\text{Orig}} + \text{M}_{\text{Base}}$}.
Notably, the adversarial error rate for \textit{xe} also dropped from 85\% to 27\% after finetuning with PTP, a decrease not observed after vanilla finetuning.
These findings suggest that targeting LLM neopronoun proficiency significantly reduces the LLM's tendency to misgender, with pronoun tokenization parity showing promise in addressing these challenges.


\subsection{Lexical Layer Finetuning}


We report results for lexical finetuning variants in \autoref{tbl:70m_model_comparison}. \textsc{$\text{T}_{\text{Orig}}$}+ \textsc{\textsc{$\text{M}_{\text{Lex}}$}} improved neopronoun consistency (16.8\%) over \textsc{$\text{T}_{\text{Orig}}$}+ \textsc{\textsc{$\text{M}_{\text{Base}}$}} and \textsc{$\text{T}_{\text{Orig}}$}+ \textsc{\textsc{$\text{M}_{\text{Full}}$}}, indicating that employing pre-existing LLM knowledge may improve neopronoun proficiency.
While lexical finetuning alone contributed modest improvements over \textsc{$\text{T}_{\text{Orig}}$}+ \textsc{\textsc{$\text{M}_{\text{Full}}$}}, \textbf{pairing lexical finetuning with PTP significantly outperformed all other models}, at 53.6\% neopronoun consistency.  
This cumulative gain, accompanied by a simultaneous reduction in adversarial error over \textsc{$\text{T}_{\text{Orig}}$}+ \textsc{$\text{M}_{\text{Full}}$} (34.8\% vs. 89.5\%), suggests a favorable synergy towards improving neopronoun morphosyntax.
We also observed gains over \textsc{$\text{T}_{\text{PTP}}$}+ \textsc{$\text{M}_{\text{Full}}$} across all data resource levels, especially at 10\% and 20\%, demonstrating its efficacy in more real-world, lower-resourced settings (further details found in Appendix~\ref{sec:appendix_size_compare}).

The impact of lexical finetuning on binary pronouns varied across models of this size. We observed stable consistency for feminine pronouns, while this was more evident for masculine pronouns with \textsc{$\text{T}_{\text{PTP}}$}+ \textsc{$\text{M}_{\text{Full}}$}. The decline in masculine pronouns after lexical training may be attributed to the distinct challenges associated with finetuning existing pronouns compared to new or under-resourced pronouns. Neopronoun tokens, which are not initialized from a pre-existing "pronoun" space, must be learned from scratch. Meanwhile, binary pronoun tokens have already converged to a meaningful lexical space. As a result, while the LLM learns these new neopronouns, the previously trained binary pronouns may be inadvertently affected. In this work, we consider it an acceptable tradeoff as it substantially improves the most disadvantaged group (i.e., equity) without severely compromising overall performance. This phenomenon is typical in bias mitigation efforts, where gains in fairness are typically balanced against performance loss. Ultimately, the optimal tradeoff is stakeholder-dependent. Future studies can build upon these findings to investigate balancing equity with overall performance further.

\subsection{Model Size Ablations}
\label{sec:ablations}

Results for all model sizes are provided in \autoref{fig:all_model_size}. Neopronoun consistency gains with PTP over finetuning with BPE tokenization were sustained across model sizes. Both \textsc{$\text{T}_{\text{PTP}} +\text{M}_{\text{Full}}$} and \textsc{$\text{T}_{\text{PTP}} +\text{M}_{\text{Lex}}$} again outperformed neopronoun consistency baselines \textsc{$\text{T}_{\text{Orig}} +\text{M}_{\text{Full}}$} and  \textsc{$\text{T}_{\text{Orig}}$}+ \textsc{\textsc{$\text{M}_{\text{Base}}$}}. Lexical finetuning performed best when paired with PTP, as found in the previous section. Across size, we also found lexical finetuning reduced compute time by up to 21.5\% over standard full finetuning (more results in Appendix~\ref{sec:hardware}).

\textsc{$\text{T}_{\text{PTP}} +\text{M}_{\text{Lex}}$} provided gains over \textsc{$\text{T}_{\text{Orig}} +\text{M}_{\text{Full}}$} across all model sizes, with larger models (>160M) benefiting most from \textsc{$\text{T}_{\text{PTP}} +\text{M}_{\text{Full}}$}. Notably, a larger model did not always improve neopronoun consistency across respective finetuning regimes. In fact, when employing PTP, \textbf{smaller models actually achieved neopronoun consistency comparable to models more than twice their size}. 
As shown in \autoref{fig:all_model_size}, a 410M model finetuned with \textsc{$\text{T}_{\text{PTP}} +\text{M}_{\text{Full}}$} resulted in the best neopronoun consistency (56.2\%), while a 160M model finetuned with \textsc{$\text{T}_{\text{PTP}} +\text{M}_{\text{Lex}}$} closely followed (53.6\%) (further details in Appendix~\ref{sec:appendix_size_compare}). 
Further examining model generations, we provide examples in \autoref{tbl:generations} which demonstrate consistent textual coherence for each of our finetuning paradigms.

\begin{table}
\footnotesize
\centering
\begin{tabular}{lp{0.72\linewidth}@{}}
\toprule
\textbf{Model} & \textbf{Text} \\
\midrule
\begin{tabular}[c]{@{}c@{}}\textbf{{$\text{T}_{\text{Orig}}+$}}\\\textbf{$\text{M}_{\text{Base}}$}\end{tabular} & \textit{Skyler's pronouns are xe/xem/xir/xirself. Xe needs to go to the bookstore soon because} it's not in the store. Xir is the same as Xirself. \\
\midrule
\begin{tabular}[c]{@{}c@{}}\textbf{{$\text{T}_{\text{Orig}}+$}}\\\textbf{$\text{M}_{\text{Full}}$}\end{tabular} & \textit{Skyler's pronouns are xe/xem/xir/xirself. Xe needs to go to the bookstore soon because} xe's too busy to be on the air. Xe's also a good friend of xir. \\
\midrule
\begin{tabular}[c]{@{}c@{}}\textbf{{$\text{T}_{\text{PTP}}+$}}\\\textbf{$\text{M}_{\text{Full}}$}\end{tabular} & \textit{Skyler's pronouns are xe/xem/xir/xirself. Xe needs to go to the bookstore soon because} xe's going to be writing a book about how to say xir name. Xe also has a book in the works called ``the art of being a writer.'' \\
\midrule
\begin{tabular}[c]{@{}c@{}}\textbf{{$\text{T}_{\text{Orig}}+$}}\\\textbf{$\text{M}_{\text{Lex}}$}\end{tabular}& \textit{Skyler's pronouns are xe/xem/xir/xirself. Xe needs to go to the bookstore soon because} xe won't have time to go tomorrow. \\
\midrule
\begin{tabular}[c]{@{}c@{}}\textbf{{$\text{T}_{\text{PTP}}+$}}\\\textbf{$\text{M}_{\text{Lex}}$}\end{tabular} & \textit{Skyler's pronouns are xe/xem/xir/xirself. Xe needs to go to the bookstore soon because} xe is a huge fan of the book ``the secret life of the apes'' by john mccarthy. \\
\bottomrule
\end{tabular}
\caption{Pythia-410M model generations across finetuning regimes.  \textit{Italics} are input prompts and generations are performed with nucleus sampling (\textsc{top-p=0.95}, \textsc{top-k=50}).}
\label{tbl:generations}
\end{table}

\section{Conclusion}

In this work, we discover how disparate BPE tokenization across gendered pronouns, a consequence of data infrequency in training corpora, is associated with a model's degraded ability to adhere to pronoun morphosyntax. 
This deficiency is highly correlated with an LLM's propensity to misgender data-scarce neopronouns.
Parallels to low-resource multilingual NLP efforts in addressing tokenizer limitations help inform novel approaches to mitigating English neopronoun misgendering. 
We find that employing vocabulary amelioration with pronoun tokenization parity along with a monolingual twist on lexical finetuning improve LLM neopronoun consistency and grammatical proficiency over traditional finetuning settings with standard BPE tokenization. 

As BPE is just one of many subword tokenization algorithms, our work opens new avenues for exploring this phenomenon under various subword tokenization algorithms and in multilingual settings. Nonetheless, these challenges ultimately arise from larger issues surrounding data availability and limitations of greedy (i.e., context-free) tokenization techniques. Addressing these foundational issues in future work is essential for sustainably developing inclusive LLMs and preventing social harm.

\section*{Limitations and Broader Impacts}
As neopronouns continue to surface and be adopted, we highlight the importance of considering how each pronoun family operates within its language. Therefore, we show this as an end-to-end example for one pronoun family in English, \textit{xe}. Future work should also consider how respective pronoun families operate within shared LLM contextual embeddings. Furthermore, adding other metrics from existing bias benchmarks may complement our study, as we mostly rely on quantitative metrics grounded in English grammar rules to assess the quality of mitigations. 

We emphasize the importance of transparent stakeholder discourse in selecting an approach that balances pronoun consistency, error rates, and case agreement.
For instance, if stakeholders choose to address historical disparities for minority groups, they may prioritize their improvement while specifying an error tolerance for dominant groups rather than solely aiming for equal or improved performance across majority groups.


\section*{Acknowledgments}
The authors thank all reviewers and chairs for their constructive feedback. Additionally, they would like to extend their appreciation to Zachary Jaggers for insightful discussions on English linguistics.

\bibliography{anthology,custom}
\bibliographystyle{acl_natbib}

\clearpage

\appendix

\section{Appendix}
\label{sec:appendix}


\subsection{Embedding Initialization}

Upon adding a new token and creating a new $E_{\text{PTP}}$, embeddings are set to default random initialization behavior in an LLM. Being that neopronouns and binary pronouns follow the same grammar rules in English, we also investigate leveraging \textit{existing} grammatical knowledge learned by the LLM to help bootstrap the model's ability to learn to use neopronouns better. Establishing a direct mapping between binary and neopronouns across their various forms, we average the neopronoun embedding with its corresponding binary pronoun embedding for each case. This approach resembles the use of a bilingual lexicon to facilitate vocabulary alignment ~\cite{Artetxe2019BilingualLI}. 


We adopt the method of taking the mean across binary pronouns for two key reasons: to leverage the LLM's syntactic knowledge related to singular pronouns used similarly to \textit{xe} in sentences and to accommodate individuals who use neopronouns and may have historical associations with binary pronouns. This is denoted in the tables from Section \ref{sec:appendix_size_compare} as PTP-B. For future work, we encourage further exploration of methods to bootstrap these embeddings.

\subsection{Model Finetuning Details}
\label{app:model_details}

\subsubsection{Experiment 1 - Full Finetuning}

We use the \textit{deduped} versions of Pythia, which trained on the Pile after the dataset had been globally deduplicated. We confirm that our research is in line with Pythia's intended use: Given their  Apache 2.0 license, we may finetune or adapt these models.

Before tokenization, text is chunked with a 256 window size, resulting in 386,267 rows before any neopronoun augmentation. We conduct finetuning with an 80/10/10 train, validation, and test split. Each model adheres to Pythia suite configurations, including an embedding size of 512 and a vocabulary size of 50,284 (50,277 without PTP). Finetuning is done for five epochs with a batch size of 128, a learning rate of $10^{-4}$, and early stopping based on cross-entropy loss on the validation set with a patience of 2. To expedite model training, all models undergo finetuning using FP16 mixed precision and 2 gradient accumulation steps.

\subsubsection{Experiment 2 - Lexical Training}
We follow the setup from the previous experiment, but only slightly increase the learning rate to 1 x $10^3$ in order to encourage more rapid adaptation to the new vocabulary.

\subsubsection{Hardware Setup}
\label{sec:hardware}
We perform all our experiments with 8 NVIDIA A100s with 40 GiB vRAM. 

\begin{table}[H]
    \small
    \centering

    \begin{tabular}{cc}
    \toprule
    \textbf{Model Size} & \textbf{Hours} \\
    \midrule
    70M & 0.65\\
    160M & 0.74 \\
    410M & 1.2 \\
    1.4B & 1.7\\
    \bottomrule
    \end{tabular}
    \caption{Average GPU Hours For Full Finetuning}
    \label{tbl:training_time}
\end{table}

\begin{table}[H]
    \small
    \centering

    \begin{tabular}{cc}
    \toprule
    \textbf{Model Size} & \textbf{Training Time Reduction (\%)} \\
    \midrule
    70M & 18.8 \\
    160M & 21.1 \\
    410M & 16.5 \\
    1.4B & 21.5 \\
    \bottomrule
    \end{tabular}
    \caption{$\Delta$ compute time switching from standard full finetuning to lexical finetuning.}
    
    \label{tbl:training_time2}
\end{table}

\begin{table}[H]
\centering
\small

\label{tbl:model_params}
\begin{tabular}{lll} 
\toprule
\textbf{Model Size} & \textbf{\# P} & \textbf{\# Non-Embedding P}  \\ 
\midrule
70M                 & 70,426,624                    & 18,915,328                                   \\
160M                & 162,322,944                   & 85,056,000                                   \\
410M                & 405,334,016                   & 302,311,424                                  \\
1.4B                & 1,414,647,808                 & 1,208,602,624                                \\
\bottomrule
\end{tabular}
\caption{Model Parameters (P), Available on \hyperlink{https://huggingface.co/EleutherAI/pythia-12b-deduped}{HuggingFace}.}
\label{tbl:number_of_params}
\end{table}

\subsection{Details on How to Reproduce PTP}
We provide details on how to reproduce PTP in Algorithm ~\ref{app:algo}.
\begin{algorithm*}
\caption{Pronoun Tokenization Parity (PTP)}
\begin{algorithmic}[1]
\State \textbf{Input 1}: LLM model 
\State \textbf{Input 2}: LLM model's BPE tokenizer
\State \textbf{Input 3}: Defined list of neopronouns for PTP 
\State \textbf{Input 4}: Dataset augmented with neopronouns 
\State \textbf{Method}: Add special tokens for each neopronoun. Be sure to explicitly add 'Ġ' to the beginning of each token to indicate that it is a full, non-subword token space before the word, otherwise this will lead to incorrect model behavior, since a lack of 'Ġ' in BPE tokenization indicates a subword token. 
\State \textbf{Check}: Check the tokenizer is working properly by checking the tokenized neopronoun, ensuring that you see 'Ġ' in its token. For example, tokenizing \textit{xe} should result in ['Ġxe'] not ['Ġ', xe']. The latter will cause the LLM to incorrectly associate a space character with a neopronoun. This can be tested by checking next word transition probabilities from the space character.
\State Resize the LLM token embeddings to match vocabulary of tokenizer. Here is example code to do this with a model and tokenizer from \texttt{HuggingFace Transformers} Package \footnote{\textit{Note: Be very careful with spacing, make sure your tokenizer can appropriately handle adding 'Ġ' to special tokens, otherwise you will need to preprocess your text to account for the extra space caused by adding 'Ġ' to each special token.} }.
\begin{footnotesize}
\begin{verbatim} 
#declare neopronoun tokens
arr_tokens = [
    'Ġxe', 'ĠXe', 
    'Ġxem', 'ĠXem', 
    'Ġxir', 'ĠXir', 
    'Ġxirs', 'ĠXirs'
]

# add new tokens to the tokenizer, t
token_dict = {
'additional_special_tokens': arr_tokens
}
t.add_special_tokens(token_dict)

# update model, m, accordingly
m.resize_token_embeddings(len(tokenizer))
\end{verbatim}
\end{footnotesize}
\If{\texttt{Lexical Finetuning}}
    \State Freeze all parameters besides the word token embeddings. Then proceed to finetune this lexical layer.
\Else
    \State Proceed with standard full finetuning
\EndIf
\State \textbf{Return} Finetuned model, new PTP tokenizer
\State Evaluate using extended \texttt{MISGENDERED} framework
\end{algorithmic}
\label{app:algo}
\end{algorithm*}
\vfill

\begin{table*}[t]
\centering
\small
\begin{tabular}{cp{10cm}} 
\toprule
\textbf{Case}                                               & \multicolumn{1}{c}{\textbf{Template}}                                                                                  \\ 
\midrule
\multirow{3}{*}{Nominative}                                 & before \{name\} came out, \{det\} \{nom\} identified differently.                                                      \\
                                                            & \{name\} is an indian filmmaker known for directing a movie, however \{det\} \{nom\} also directs television.          \\
                                                            & \{name\} is a musician and dancer and \{det\} \{nom\} currently lives somewhere nearby.                                \\ 
\midrule
\multirow{3}{*}{Accusative}                                 & upon recognizing \{name\}, the fan asked \{det\} \{acc\} for an autograph.                                             \\
                                                            & \{name\} owns the car so be sure to give the keys back to \{det\} \{acc\} afterwards.                                  \\
                                                            & \{name\} needs new shoes so I can give my other shoes to \{det\} \{acc\}.                                              \\ 
\midrule
\multirow{3}{*}{Pronominal Possessive}                      & \{name\} was a participant in \{det\} \{gen-indep\} own musical.                                                       \\
                                                            & \{name\} is an author of children's fantasy, best known for \{det\} \{gen-indep\} book that won several awards.        \\
                                                            & \{name\} is a singer-songwriter-producer based in boston and \{det\} \{gen-indep\} debut album was released in march.  \\ 
\midrule
\multicolumn{1}{l}{\multirow{3}{*}{Predicative Possessive}} & \{name\} was working in this area, so the phone is probably \{det\} \{gen-dep\}.                                       \\
\multicolumn{1}{l}{}                                        & the story of \{name\}'s resilience and triumph over adversity is \{det\} \{gen-dep\} to own.                           \\
\multicolumn{1}{l}{}                                        & name\} said to me that the larger slice of pizza was mine, and the smaller one was \{det\} \{gen-dep\}.                \\ 
\midrule
\multirow{3}{*}{Reflexive}                                  & \{name\} will read the book by \{det\} \{reflex\}.                                                                     \\
                                                            & \{name\} needs to be by \{det\} \{reflex\} sometimes.                                                                  \\
                                                            & \{name\} often works alone by \{det\} \{reflex\}.                                                                      \\
\bottomrule
\end{tabular}
\caption{Template additions to \texttt{MISGENDERED} 
\label{tbl:template_additions}
\cite{hossain2023misgendered}}
\end{table*}

\vfill



\subsection{Templates additions to \texttt{MISGENDERED}}
\label{app:more_templates}
To mimic real world pronoun declarations, each declaration is started with nominative, accusative, pronominal possessive, and reflexive pronouns. 

\autoref{tbl:template_additions} reflects selected additions from the \texttt{TANGO} dataset. \texttt{Det} represents the determiner position one may replace with ones like \textit{the, a, these, those}. \texttt{Gen-dep, Gen-indep, reflex, nom} are all pronoun cases.

\subsection{Example Generations}
\label{app:text_generations}
\autoref{tbl:generations} example generations from the prompt \textit{Skyler's pronouns are xe/xem/xir/xirself.}

\subsection{PTP Training Evaluation}
\begin{figure*}[t]
    \centering
    \includegraphics[width=\linewidth]{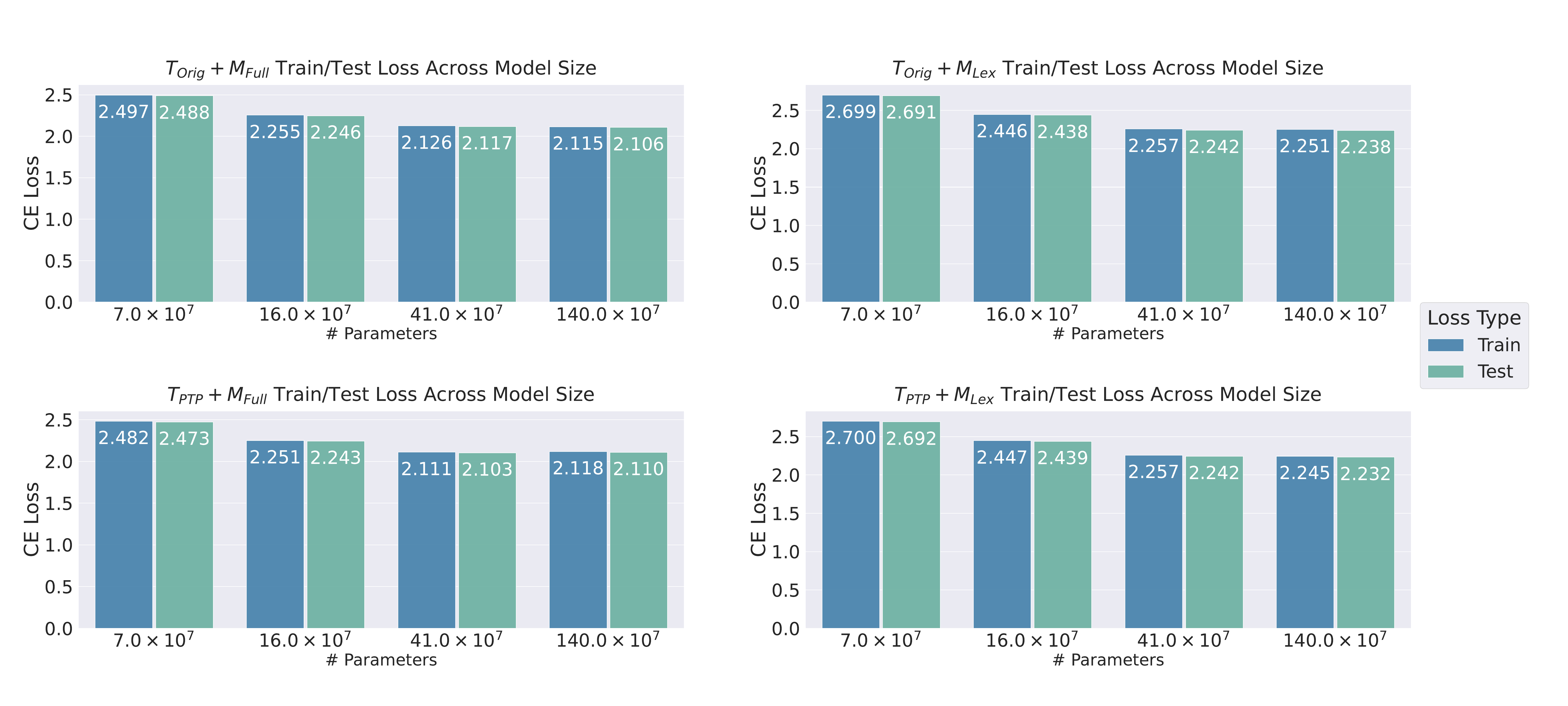}
    \vspace{-1cm}
    \caption{Reported Cross Entropy Loss for train/test across models.}
    \label{app:training_evalconf}
\end{figure*}

We report cross entropy loss for the train and test across each model in \autoref{app:training_evalconf}.

\label{sec:training_evalconf}

\subsection{Downstream Evaluations}
\label{app:evaluations}
\subsubsection{Setup}
To confirm that our proposed techniques do not adversely affect downstream performance, we assess our models on three benchmarks for pronoun resolution and coreference resolution, logical reasoning , and knowledge retrieval respectively: \textsc{winogrande} (5-shot) \citep{sakaguchi2021winogrande}, \textsc{LogiQA} (5-shot) \citep{liu2020logiqa}, and \textsc{ARC-Challenge} (5-shot) \citep{clark2018think}. We utilize the LM evaluation harness \footnote{\url{https://github.com/EleutherAI/lm-evaluation-harness}} and discuss the results in the following subsections.




\subsubsection{Results}
\begin{table}
\renewcommand{\arraystretch}{1} 
\setlength{\tabcolsep}{4pt} 
\footnotesize
\centering
\begin{tabular}{lllll} 
\toprule
\textbf{Size}                  & \textbf{Version}                                          & \textbf{Wino}                         & \textbf{ARC}                    & \textbf{LogiQA}                             \\ 
\midrule
\multirow{5}{*}{70M}  & Base                                         & 49.17\textsubscript{1.41}          & 19.71\textsubscript{1.16}          & 27.96\textsubscript{1.76}           \\
                      & $\text{T}_\text{Orig} + \text{M}_\text{Full}$    & 49.64\textsubscript{1.41}          & 22.18\textsubscript{1.21}          & 26.57\textsubscript{1.73}           \\
                      & $\text{T}_\text{PTP} + \text{M}_\text{Base}$ & 50.43\textsubscript{1.41}          & 21.93\textsubscript{1.21}          & 25.50\textsubscript{1.71}           \\
                      & $\text{T}_\text{Orig} + \text{M}_\text{Lex}$     & 50.51\textsubscript{1.41}          & \textbf{23.72\textsubscript{1.24}} & 29.03\textsubscript{1.78}           \\
                      & $\text{T}_\text{PTP} + \text{M}_\text{Lex}$      & \textbf{50.99\textsubscript{1.40}} & \textbf{23.72\textsubscript{1.24}} & \textbf{29.49\textsubscript{1.79}}  \\ 
\midrule
\multirow{5}{*}{160M} & Base                                         & 49.72\textsubscript{1.41}          & 23.63\textsubscript{1.24}          & 26.27\textsubscript{1.73}           \\
                      & $\text{T}_\text{Orig} + \text{M}_\text{Full}$    & 52.09\textsubscript{1.40}          & 24.74\textsubscript{1.26}          & 25.96\textsubscript{1.72}           \\
                      & $\text{T}_\text{PTP} + \text{M}_\text{Base}$ & \textbf{52.17\textsubscript{1.40}} & 24.15\textsubscript{1.25}          & 27.04\textsubscript{1.74}           \\
                      & $\text{T}_\text{Orig} + \text{M}_\text{Lex}$     & 48.38\textsubscript{1.40}          & 24.49\textsubscript{1.26}          & \textbf{29.80\textsubscript{1.79}}  \\
                      & $\text{T}_\text{PTP} + \text{M}_\text{Lex}$      & 48.15\textsubscript{1.40}          & \textbf{26.02\textsubscript{1.28}} & 29.49\textsubscript{1.79}           \\ 
\midrule
\multirow{5}{*}{410M} & Base                                         & \textbf{54.85\textsubscript{1.40}} & 25.85\textsubscript{1.28}          & 24.12\textsubscript{1.68}           \\
                      & $\text{T}_\text{Orig} + \text{M}_\text{Full}$    & 52.88\textsubscript{1.40}          & 27.22\textsubscript{1.30}          & 27.04\textsubscript{1.74}           \\
                      & $\text{T}_\text{PTP} + \text{M}_\text{Base}$ & 52.96\textsubscript{1.40}          & 25.34\textsubscript{1.27}          & 26.57\textsubscript{1.73}           \\
                      & $\text{T}_\text{Orig} + \text{M}_\text{Lex}$     & 51.46\textsubscript{1.40}          & 25.26\textsubscript{1.27}          & \textbf{27.80\textsubscript{1.76}}  \\
                      & $\text{T}_\text{PTP} + \text{M}_\text{Lex}$      & 51.46\textsubscript{1.40}          & \textbf{27.39\textsubscript{1.30}} & 27.50\textsubscript{1.75}           \\ 
\midrule
\multirow{5}{*}{1.4B} & Base                                         & \textbf{56.43\textsubscript{1.39}} & \textbf{32.17\textsubscript{1.37}} & 22.89\textsubscript{1.65}           \\
                      & $\text{T}_\text{Orig} + \text{M}_\text{Full}$    & 53.75\textsubscript{1.40}          & 26.19\textsubscript{1.28}          & 27.34\textsubscript{1.75}           \\
                      & $\text{T}_\text{PTP} + \text{M}_\text{Base}$ & 53.99\textsubscript{1.40}          & 24.91\textsubscript{1.26}          & 26.88\textsubscript{1.74}           \\
                      & $\text{T}_\text{Orig} + \text{M}_\text{Lex}$     & 52.72\textsubscript{1.40}          & 30.03\textsubscript{1.34}          & \textbf{28.73\textsubscript{1.77}}  \\
                      & $\text{T}_\text{PTP} + \text{M}_\text{Lex}$      & 52.72\textsubscript{1.40}          & 28.75\textsubscript{1.32}          & 28.57\textsubscript{1.77}           \\
\bottomrule
\end{tabular}
\caption{Downstream Evaluations Across Model Size. Subscripts reflect standard deviations.}
\label{tbl:bench_paper}
\end{table}
%

We report our results in \autoref{tbl:bench_paper}. For Winogrande, half of the models employing our methods either sustain or slightly boost performance, ranging from 0.08 to 0.24 points, likely due to improvement in pronoun disambiguation. For 410M and 1.4B, this boost is not observed. These base models slightly outperform our experiments, though the differences are marginal (1-2\%) and insignificant.

For ARC, PTP and lexical finetuning either sustain or slightly improve baseline performance  (1-2\%) for most model sizes. For the 70M model, all lexical training outperforms full finetuning with original tokenization and the base model. We find this pattern consistent for 160M, and 410M. For the 1.4B model, the base model outperforms regular full finetuning with a 7\% gap for full finetuning on original tokenization. In contrast, both lexical techniques outperform finetuning with both the original tokenizer and PTP. This finding indicates that combining PTP with lexical layer finetuning may be the best option for the highest pronoun gains while maintaining existing LLM capabilities.

For LogiQA, our methods either improve or are within the range of the baseline model. Namely, lexical finetuning corresponds to a good improvement over baseline. This finding is likely related to focused improvements in the LLM's lexical layers overall. Across all model sizes, both lexical training consistently outperforms finetuning without PTP and the base models. Our findings suggest that lexical layer finetuning, with or without vocabulary expansion, does not harm the model's downstream performance on LogiQA compared to regular finetuning or the base models.

\subsection{Ablations}
\label{sec:appendix_size_compare}


\begin{table*}[t]
\tiny
\centering

\begin{subtable}{\textwidth}
\centering
\begin{tabu}{l|ccc|lll|lll} 
\toprule
\multicolumn{1}{c|}{\multirow{2}{*}{\textbf{Model}}}                             & \multicolumn{3}{c|}{\textbf{Consistency ($\uparrow$)}}                                                       & \multicolumn{3}{c|}{\textbf{Case Error ($\downarrow$)}}                                                     & \multicolumn{3}{c}{\textbf{Inject Error ($\downarrow$)}}                                                      \\
\multicolumn{1}{c|}{}                                                            & \textbf{He}                        & \textbf{She}                       & \textbf{Xe}                        & \multicolumn{1}{c}{\textbf{He}}   & \multicolumn{1}{c}{\textbf{She}}   & \multicolumn{1}{c|}{\textbf{Xe}}   & \multicolumn{1}{c}{\textbf{He}}    & \multicolumn{1}{c}{\textbf{She}}   & \multicolumn{1}{c}{\textbf{Xe}}     \\ 
\midrule
\rowfont{\color{gray}}\textsc{$\text{T}_{\text{Orig}} + \text{M}_{\text{Base}}$} & \textbf{96.82\textsubscript{0.79}} & 71.59\textsubscript{2.00}          & 0.67\textsubscript{0.33}           & \textbf{8.26\textsubscript{1.23}} & \textbf{24.36\textsubscript{1.92}} & 78.56\textsubscript{1.82}          & \textbf{23.85\textsubscript{1.87}} & 16.92\textsubscript{1.67}          & 85.03\textsubscript{1.59}           \\
\textsc{$\text{T}_{\text{Orig}} + \text{M}_{\text{Full}}$}                       & 89.64\textsubscript{1.33}          & 86.05\textsubscript{1.54}          & 14.46\textsubscript{1.56}          & 11.74\textsubscript{1.44}         & 22.41\textsubscript{1.82}          & 59.95\textsubscript{2.18}          & 23.95\textsubscript{1.90}          & \textbf{16.77\textsubscript{1.69}} & 89.49\textsubscript{1.36}           \\
\textsc{$\text{T}_{\text{Orig}} + \text{M}_{\text{Lex}}$}                        & 86.46\textsubscript{1.54}          & 72.87\textsubscript{1.97}          & 16.77\textsubscript{1.67}          & 18.51\textsubscript{1.74}         & 33.79\textsubscript{2.10}          & 70.51\textsubscript{2.00}          & 28.97\textsubscript{2.03}          & 23.18\textsubscript{1.87}          & 65.44\textsubscript{2.13}           \\
\textsc{$\text{T}_{\text{PTP}} + \text{M}_{\text{Full}}$}                        & 94.77\textsubscript{0.97}          & \textbf{83.49\textsubscript{1.64}} & 37.79\textsubscript{2.18}          & 9.69\textsubscript{1.31}          & 29.28\textsubscript{2.00}          & \textbf{56.92\textsubscript{2.21}} & 27.79\textsubscript{1.97}          & 20.97\textsubscript{1.82}          & 27.03\textsubscript{1.92}           \\
\textsc{$\text{T}_{\text{PTP-B}} + \text{M}_{\text{Full}}$}                      & 96.21\textsubscript{0.85}          & 80.72\textsubscript{1.77}          & 24.36\textsubscript{1.90}          & 9.49\textsubscript{1.31}          & 31.33\textsubscript{2.05}          & 61.90\textsubscript{2.18}          & 28.26\textsubscript{2.03}          & 20.56\textsubscript{1.77}          & \textbf{25.95\textsubscript{1.95}}  \\
\textsc{$\text{T}_{\text{PTP}} + \text{M}_{\text{Lex}}$}                         & 84.97\textsubscript{1.56}          & 72.21\textsubscript{1.97}          & \textbf{53.59\textsubscript{2.23}} & 18.15\textsubscript{1.69}         & 33.03\textsubscript{2.10}          & 60.46\textsubscript{2.15}          & 25.79\textsubscript{1.95}          & 21.85\textsubscript{1.85}          & 34.77\textsubscript{2.10}           \\
\textsc{$\text{T}_{\text{PTP-B}} + \text{M}_{\text{Lex}}$}                       & 83.28\textsubscript{1.64}          & 74.31\textsubscript{1.97}          & 42.97\textsubscript{2.23}          & 16.10\textsubscript{1.64}         & 33.33\textsubscript{2.08}          & 57.74\textsubscript{2.18}          & 24.31\textsubscript{1.90}          & 20.21\textsubscript{1.79}          & 32.05\textsubscript{2.08}           \\
\bottomrule
\end{tabu}
\caption{Data Split=10}
\end{subtable}

\begin{subtable}{\textwidth}
\centering
\begin{tabu}{l|ccc|lll|lll} 
\toprule
\multicolumn{1}{c|}{\multirow{2}{*}{\textbf{Model}}}                                                & \multicolumn{3}{c|}{\textbf{Consistency ($\uparrow$)}}                                                       & \multicolumn{3}{c|}{\textbf{Case Error ($\downarrow$)}}                                                     & \multicolumn{3}{c}{\textbf{Inject Error ($\downarrow$)}}                                                      \\
                                                                                 & \textbf{He}                        & \textbf{She}                       & \textbf{Xe}                        & \multicolumn{1}{c}{\textbf{He}}   & \multicolumn{1}{c}{\textbf{She}}   & \multicolumn{1}{c|}{\textbf{Xe}}   & \multicolumn{1}{c}{\textbf{He}}    & \multicolumn{1}{c}{\textbf{She}}   & \multicolumn{1}{c}{\textbf{Xe}}     \\ 
\midrule
\rowfont{\color{gray}}\textsc{$\text{T}_{\text{Orig}} + \text{M}_{\text{Base}}$} & \textbf{96.82\textsubscript{0.77}} & 71.59\textsubscript{2.03}          & 0.67\textsubscript{0.36}           & \textbf{8.26\textsubscript{1.23}} & \textbf{24.36\textsubscript{1.90}} & 78.56\textsubscript{1.85}          & \textbf{23.85\textsubscript{1.87}} & \textbf{16.92\textsubscript{1.67}} & 85.03\textsubscript{1.59}           \\
\textsc{$\text{T}_{\text{Orig}} + \text{M}_{\text{Full}}$}                       & 93.23\textsubscript{1.10}          & 81.44\textsubscript{1.77}          & 13.74\textsubscript{1.56}          & 11.69\textsubscript{1.44}         & 24.97\textsubscript{1.92}          & 58.77\textsubscript{2.15}          & 27.08\textsubscript{1.95}          & 18.00\textsubscript{1.72}          & 87.28\textsubscript{1.46}           \\
\textsc{$\text{T}_{\text{Orig}} + \text{M}_{\text{Lex}}$}                        & 86.05\textsubscript{1.51}          & 73.33\textsubscript{1.95}          & 18.10\textsubscript{1.69}          & 17.03\textsubscript{1.67}         & 32.00\textsubscript{2.10}          & 71.38\textsubscript{2.00}          & 27.59\textsubscript{1.97}          & 19.90\textsubscript{1.74}          & 67.54\textsubscript{2.10}           \\
\textsc{$\text{T}_{\text{PTP}} + \text{M}_{\text{Full}}$}                        & 96.51\textsubscript{0.82}          & \textbf{88.56\textsubscript{1.38}} & 35.95\textsubscript{2.10}          & 11.59\textsubscript{1.41}         & 32.05\textsubscript{2.05}          & \textbf{47.54\textsubscript{2.21}} & 25.28\textsubscript{1.97}          & 19.18\textsubscript{1.77}          & \textbf{33.85\textsubscript{2.05}}  \\
\textsc{$\text{T}_{\text{PTP-B}} + \text{M}_{\text{Full}}$}                      & 95.28\textsubscript{0.92}          & 87.33\textsubscript{1.46}          & 18.51\textsubscript{1.69}          & 9.95\textsubscript{1.33}          & 30.72\textsubscript{2.00}          & 48.41\textsubscript{2.18}          & 26.87\textsubscript{1.92}          & 19.54\textsubscript{1.74}          & 34.00\textsubscript{2.10}           \\
\textsc{$\text{T}_{\text{PTP}} + \text{M}_{\text{Lex}}$}                         & 82.21\textsubscript{1.69}          & 70.87\textsubscript{2.03}          & \textbf{48.00\textsubscript{2.23}} & 15.44\textsubscript{1.64}         & 31.59\textsubscript{2.05}          & 59.23\textsubscript{2.18}          & 30.10\textsubscript{2.03}          & 23.69\textsubscript{1.87}          & 34.92\textsubscript{2.10}           \\
\textsc{$\text{T}_{\text{PTP-B}} + \text{M}_{\text{Lex}}$}                       & 83.18\textsubscript{1.67}          & 70.05\textsubscript{2.00}          & 32.41\textsubscript{2.03}          & 15.28\textsubscript{1.59}         & 32.92\textsubscript{2.08}          & 57.95\textsubscript{2.21}          & 30.05\textsubscript{2.05}          & 22.62\textsubscript{1.87}          & 34.00\textsubscript{2.08}           \\
\bottomrule
\end{tabu}
\caption{Data Split=20}
\end{subtable}

\begin{subtable}{\textwidth}
\centering
\begin{tabu}{l|ccc|lll|lll} 
\toprule
\multicolumn{1}{c|}{\multirow{2}{*}{\textbf{Model}}}                                                    & \multicolumn{3}{c|}{\textbf{Consistency ($\uparrow$)}}                                                       & \multicolumn{3}{c|}{\textbf{Case Error ($\downarrow$)}}                                                     & \multicolumn{3}{c}{\textbf{Inject Error ($\downarrow$)}}                                                      \\
                                                                                 & \textbf{He}                        & \textbf{She}                       & \textbf{Xe}                        & \multicolumn{1}{c}{\textbf{He}}   & \multicolumn{1}{c}{\textbf{She}}   & \multicolumn{1}{c|}{\textbf{Xe}}   & \multicolumn{1}{c}{\textbf{He}}    & \multicolumn{1}{c}{\textbf{She}}   & \multicolumn{1}{c}{\textbf{Xe}}     \\ 
\midrule
\rowfont{\color{gray}}\textsc{$\text{T}_{\text{Orig}} + \text{M}_{\text{Base}}$} & \textbf{96.82\textsubscript{0.77}} & 71.59\textsubscript{2.00}          & 0.67\textsubscript{0.36}           & \textbf{8.26\textsubscript{1.23}} & 24.36\textsubscript{1.90}          & 78.56\textsubscript{1.82}          & \textbf{23.85\textsubscript{1.87}} & \textbf{16.92\textsubscript{1.64}} & 85.03\textsubscript{1.59}           \\
\textsc{$\text{T}_{\text{Orig}} + \text{M}_{\text{Full}}$}                       & 91.28\textsubscript{1.26}          & \textbf{85.23\textsubscript{1.59}} & 13.85\textsubscript{1.51}          & 12.87\textsubscript{1.46}         & \textbf{21.90\textsubscript{1.82}} & 60.62\textsubscript{2.18}          & 24.56\textsubscript{1.90}          & 19.08\textsubscript{1.72}          & 87.03\textsubscript{1.49}           \\
\textsc{$\text{T}_{\text{Orig}} + \text{M}_{\text{Lex}}$}                        & 80.10\textsubscript{1.77}          & 64.67\textsubscript{2.10}          & 18.46\textsubscript{1.74}          & 22.62\textsubscript{1.85}         & 34.56\textsubscript{2.10}          & 68.87\textsubscript{2.08}          & 29.18\textsubscript{2.00}          & 24.26\textsubscript{1.92}          & 66.56\textsubscript{2.10}           \\
\textsc{$\text{T}_{\text{PTP}} + \text{M}_{\text{Full}}$}                        & 95.79\textsubscript{0.90}          & 87.69\textsubscript{1.44}          & 32.41\textsubscript{2.08}          & 13.44\textsubscript{1.51}         & 28.51\textsubscript{2.00}          & \textbf{46.92\textsubscript{2.18}} & 23.18\textsubscript{1.90}          & 19.69\textsubscript{1.74}          & 34.41\textsubscript{2.13}           \\
\textsc{$\text{T}_{\text{PTP-B}} + \text{M}_{\text{Full}}$}                      & 90.87\textsubscript{1.28}          & 84.41\textsubscript{1.56}          & 12.56\textsubscript{1.49}          & 10.46\textsubscript{1.36}         & 30.00\textsubscript{2.05}          & 49.33\textsubscript{2.23}          & 25.49\textsubscript{1.95}          & 19.13\textsubscript{1.74}          & \textbf{26.00\textsubscript{1.97}}  \\
\textsc{$\text{T}_{\text{PTP}} + \text{M}_{\text{Lex}}$}                         & 81.23\textsubscript{1.72}          & 62.00\textsubscript{2.15}          & 45.49\textsubscript{2.21}          & 19.64\textsubscript{1.77}         & 35.74\textsubscript{2.15}          & 55.49\textsubscript{2.18}          & 26.77\textsubscript{1.95}          & 20.92\textsubscript{1.82}          & 31.44\textsubscript{2.05}           \\
\textsc{$\text{T}_{\text{PTP-B}} + \text{M}_{\text{Lex}}$}                       & 84.87\textsubscript{1.59}          & 69.33\textsubscript{2.08}          & \textbf{48.26\textsubscript{2.23}} & 20.72\textsubscript{1.79}         & 35.79\textsubscript{2.08}          & 53.33\textsubscript{2.23}          & 27.69\textsubscript{2.00}          & 20.97\textsubscript{1.79}          & 33.33\textsubscript{2.10}           \\
\bottomrule
\end{tabu}
\caption{Data Split=30}
\end{subtable}

\begin{subtable}{\textwidth}
\centering
\begin{tabu}{l|ccc|lll|lll} 
\toprule
\multicolumn{1}{c|}{\multirow{2}{*}{\textbf{Model}}}                                                    & \multicolumn{3}{c|}{\textbf{Consistency ($\uparrow$)}}                                                       & \multicolumn{3}{c|}{\textbf{Case Error ($\downarrow$)}}                                                     & \multicolumn{3}{c}{\textbf{Inject Error ($\downarrow$)}}                                                      \\
                                                                                 & \textbf{He}                        & \textbf{She}                       & \textbf{Xe}                        & \textbf{He}                       & \textbf{She}                       & \textbf{Xe}                        & \textbf{He}                        & \textbf{She}                       & \textbf{Xe}                         \\ 
\midrule
\rowfont{\color{gray}}\textsc{$\text{T}_{\text{Orig}} + \text{M}_{\text{Base}}$} & \textbf{96.82\textsubscript{0.77}} & 71.59\textsubscript{2.00}          & 0.67\textsubscript{0.38}           & \textbf{8.26\textsubscript{1.23}} & 24.36\textsubscript{1.87}          & 78.56\textsubscript{1.79}          & \textbf{23.85\textsubscript{1.87}} & \textbf{16.92\textsubscript{1.67}} & 85.03\textsubscript{1.59}           \\
\textsc{$\text{T}_{\text{Orig}} + \text{M}_{\text{Full}}$}                       & 92.41\textsubscript{1.18}          & 79.33\textsubscript{1.79}          & 9.95\textsubscript{1.31}           & 14.97\textsubscript{1.56}         & \textbf{21.28\textsubscript{1.79}} & 60.31\textsubscript{2.15}          & 25.38\textsubscript{1.92}          & 19.64\textsubscript{1.79}          & 85.54\textsubscript{1.56}           \\
\textsc{$\text{T}_{\text{Orig}} + \text{M}_{\text{Lex}}$}                        & 82.15\textsubscript{1.72}          & 65.23\textsubscript{2.15}          & 18.21\textsubscript{1.72}          & 24.00\textsubscript{1.92}         & 33.03\textsubscript{2.05}          & 67.38\textsubscript{2.08}          & 31.08\textsubscript{2.08}          & 22.56\textsubscript{1.85}          & 68.15\textsubscript{2.08}           \\
\textsc{$\text{T}_{\text{PTP}} + \text{M}_{\text{Full}}$}                        & 96.00\textsubscript{0.87}          & \textbf{86.41\textsubscript{1.56}} & 26.82\textsubscript{1.95}          & 15.33\textsubscript{1.59}         & 32.77\textsubscript{2.08}          & \textbf{47.38\textsubscript{2.21}} & 25.13\textsubscript{1.95}          & 20.00\textsubscript{1.72}          & 33.90\textsubscript{2.13}           \\
\textsc{$\text{T}_{\text{PTP-B}} + \text{M}_{\text{Full}}$}                      & 96.67\textsubscript{0.79}          & 86.15\textsubscript{1.49}          & 11.69\textsubscript{1.44}          & 8.72\textsubscript{1.23}          & 32.00\textsubscript{2.05}          & 48.21\textsubscript{2.23}          & 23.44\textsubscript{1.85}          & 20.26\textsubscript{1.77}          & 33.95\textsubscript{2.10}           \\
\textsc{$\text{T}_{\text{PTP}} + \text{M}_{\text{Lex}}$}                         & 85.33\textsubscript{1.56}          & 61.49\textsubscript{2.15}          & \textbf{48.41\textsubscript{2.26}} & 22.15\textsubscript{1.85}         & 37.74\textsubscript{2.13}          & 53.59\textsubscript{2.15}          & 28.97\textsubscript{2.03}          & 21.64\textsubscript{1.79}          & 33.18\textsubscript{2.10}           \\
\textsc{$\text{T}_{\text{PTP-B}} + \text{M}_{\text{Lex}}$}                       & 84.92\textsubscript{1.59}          & 62.00\textsubscript{2.21}          & 41.44\textsubscript{2.21}          & 21.69\textsubscript{1.82}         & 38.26\textsubscript{2.15}          & 53.08\textsubscript{2.21}          & 28.92\textsubscript{2.00}          & 22.87\textsubscript{1.87}          & \textbf{33.08\textsubscript{2.10}}  \\
\bottomrule
\end{tabu}
\caption{Data Split=40}
\end{subtable}

\begin{subtable}{\textwidth}
\centering
\begin{tabu}{l|ccc|lll|lll} 
\toprule
\multicolumn{1}{c|}{\multirow{2}{*}{\textbf{Model}}}                                                   & \multicolumn{3}{c|}{\textbf{Consistency ($\uparrow$)}}                                                       & \multicolumn{3}{c|}{\textbf{Case Error ($\downarrow$)}}                                                     & \multicolumn{3}{c}{\textbf{Inject Error ($\downarrow$)}}                                                      \\
                                                                                 & \textbf{He}                        & \textbf{She}                       & \textbf{Xe}                        & \multicolumn{1}{c}{\textbf{He}}   & \multicolumn{1}{c}{\textbf{She}}   & \multicolumn{1}{c|}{\textbf{Xe}}   & \multicolumn{1}{c}{\textbf{He}}    & \multicolumn{1}{c}{\textbf{She}}   & \multicolumn{1}{c}{\textbf{Xe}}     \\ 
\midrule
\rowfont{\color{gray}}\textsc{$\text{T}_{\text{Orig}} + \text{M}_{\text{Base}}$} & \textbf{96.82\textsubscript{0.79}} & 71.59\textsubscript{2.00}          & 0.67\textsubscript{0.36}           & \textbf{8.26\textsubscript{1.23}} & 24.36\textsubscript{1.87}          & 78.56\textsubscript{1.85}          & \textbf{23.85\textsubscript{1.90}} & \textbf{16.92\textsubscript{1.64}} & 85.03\textsubscript{1.56}           \\
\textsc{$\text{T}_{\text{Orig}} + \text{M}_{\text{Full}}$}                       & 93.44\textsubscript{1.08}          & 85.23\textsubscript{1.54}          & 14.05\textsubscript{1.54}          & 9.59\textsubscript{1.33}          & \textbf{23.08\textsubscript{1.87}} & 59.28\textsubscript{2.18}          & 26.00\textsubscript{1.97}          & 19.79\textsubscript{1.79}          & 86.10\textsubscript{1.54}           \\
\textsc{$\text{T}_{\text{Orig}} + \text{M}_{\text{Lex}}$}                        & 83.38\textsubscript{1.67}          & 65.13\textsubscript{2.13}          & 18.46\textsubscript{1.69}          & 20.51\textsubscript{1.79}         & 36.82\textsubscript{2.13}          & 69.54\textsubscript{2.05}          & 28.72\textsubscript{2.05}          & 19.03\textsubscript{1.72}          & 71.18\textsubscript{2.03}           \\
\textsc{$\text{T}_{\text{PTP}} + \text{M}_{\text{Full}}$}                        & 96.00\textsubscript{0.87}          & \textbf{88.92\textsubscript{1.36}} & 26.67\textsubscript{1.97}          & 13.64\textsubscript{1.54}         & 31.64\textsubscript{2.08}          & \textbf{45.90\textsubscript{2.23}} & 24.36\textsubscript{1.90}          & 21.69\textsubscript{1.87}          & 35.90\textsubscript{2.10}           \\
\textsc{$\text{T}_{\text{PTP-B}} + \text{M}_\text{Full}$}                        & 95.03\textsubscript{0.97}          & 87.23\textsubscript{1.51}          & 16.10\textsubscript{1.64}          & 10.97\textsubscript{1.38}         & 33.08\textsubscript{2.05}          & 48.36\textsubscript{2.21}          & 29.49\textsubscript{2.03}          & 21.59\textsubscript{1.82}          & 37.90\textsubscript{2.15}           \\
\textsc{$\text{T}_{\text{PTP}} + \text{M}_{\text{Lex}}$}                         & 77.54\textsubscript{1.85}          & 58.15\textsubscript{2.23}          & \textbf{58.41\textsubscript{2.18}} & 21.64\textsubscript{1.85}         & 37.74\textsubscript{2.18}          & 50.87\textsubscript{2.23}          & 29.13\textsubscript{2.03}          & 19.74\textsubscript{1.82}          & \textbf{31.54\textsubscript{2.03}}  \\
\textsc{$\text{T}_{\text{PTP-B}} + \text{M}_{\text{Lex}}$}                       & 81.54\textsubscript{1.72}          & 64.41\textsubscript{2.13}          & 49.28\textsubscript{2.21}          & 19.95\textsubscript{1.77}         & 37.85\textsubscript{2.13}          & 52.67\textsubscript{2.21}          & 26.77\textsubscript{1.92}          & 22.41\textsubscript{1.82}          & 30.51\textsubscript{2.05}           \\
\bottomrule
\end{tabu}
\caption{Data Split=50}
\end{subtable}
\caption{70M Model Results Across Data Splits}
\label{app:70m_compare}

\end{table*}

\begin{table*}[t]
\tiny
\begin{subtable}{\textwidth}
\centering
\begin{tabu}{l|ccc|lll|lll} 
\toprule
\multicolumn{1}{c|}{\multirow{2}{*}{\textbf{Model}}}                                                    & \multicolumn{3}{c|}{\textbf{Consistency ($\uparrow$)}}                                                       & \multicolumn{3}{c|}{\textbf{Case Error ($\downarrow$)}}                                                     & \multicolumn{3}{c}{\textbf{Inject Error ($\downarrow$)}}                                                     \\
                                                                                 & \textbf{He}                        & \textbf{She}                       & \textbf{Xe}                        & \multicolumn{1}{c}{\textbf{He}}   & \multicolumn{1}{c}{\textbf{She}}   & \multicolumn{1}{c|}{\textbf{Xe}}   & \multicolumn{1}{c}{\textbf{He}}   & \multicolumn{1}{c}{\textbf{She}}   & \multicolumn{1}{c}{\textbf{Xe}}     \\ 
\midrule
\rowfont{\color{gray}}\textsc{$\text{T}_{\text{Orig}} + \text{M}_{\text{Base}}$} & 79.95\textsubscript{1.77}          & \textbf{76.46\textsubscript{1.87}} & 0.00\textsubscript{0.00}           & \textbf{4.05\textsubscript{0.85}} & \textbf{10.87\textsubscript{1.36}} & 80.00\textsubscript{1.74}          & \textbf{8.72\textsubscript{1.26}} & 6.46\textsubscript{1.08}           & 95.38\textsubscript{0.95}           \\
\textsc{$\text{T}_{\text{Orig}} + \text{M}_{\text{Full}}$}                       & 78.87\textsubscript{1.79}          & 61.49\textsubscript{2.15}          & 15.59\textsubscript{1.64}          & 11.23\textsubscript{1.38}         & 20.21\textsubscript{1.77}          & 48.92\textsubscript{2.21}          & 19.44\textsubscript{1.74}         & 20.31\textsubscript{1.79}          & 69.18\textsubscript{2.03}           \\
\textsc{$\text{T}_{\text{Orig}} + \text{M}_{\text{Lex}}$}                        & 77.28\textsubscript{1.82}          & 70.05\textsubscript{2.03}          & 20.00\textsubscript{1.77}          & 12.56\textsubscript{1.46}         & 23.90\textsubscript{1.90}          & 57.59\textsubscript{2.18}          & 20.21\textsubscript{1.79}         & 16.87\textsubscript{1.67}          & 78.26\textsubscript{1.85}           \\
\textsc{$\text{T}_{\text{PTP}} + \text{M}_{\text{Full}}$}                        & 80.21\textsubscript{1.79}          & 64.92\textsubscript{2.08}          & 30.92\textsubscript{2.03}          & 6.21\textsubscript{1.05}          & 23.59\textsubscript{1.90}          & 56.26\textsubscript{2.18}          & 22.00\textsubscript{1.87}         & 18.15\textsubscript{1.72}          & \textbf{14.72\textsubscript{1.62}}  \\
\textsc{$\text{T}_{\text{PTP-B}} + \text{M}_{\text{Full}}$}                      & 79.13\textsubscript{1.79}          & 65.79\textsubscript{2.08}          & 9.74\textsubscript{1.33}           & 8.26\textsubscript{1.21}          & 22.51\textsubscript{1.87}          & 59.85\textsubscript{2.15}          & 20.87\textsubscript{1.79}         & 21.03\textsubscript{1.82}          & 25.28\textsubscript{1.92}           \\
\textsc{$\text{T}_{\text{PTP}} + \text{M}_{\text{Lex}}$}                         & 78.51\textsubscript{1.82}          & 60.77\textsubscript{2.21}          & 51.79\textsubscript{2.23}          & 12.10\textsubscript{1.41}         & 27.64\textsubscript{1.97}          & \textbf{46.36\textsubscript{2.23}} & 19.13\textsubscript{1.74}         & \textbf{14.77\textsubscript{1.62}} & 31.44\textsubscript{2.03}           \\
\textsc{$\text{T}_{\text{PTP-B}} + \text{M}_{\text{Lex}}$}                       & \textbf{81.23\textsubscript{1.72}} & 60.46\textsubscript{2.15}          & \textbf{53.64\textsubscript{2.18}} & 13.38\textsubscript{1.51}         & 29.18\textsubscript{2.00}          & 47.49\textsubscript{2.21}          & 17.49\textsubscript{1.69}         & 16.41\textsubscript{1.67}          & 25.13\textsubscript{1.95}           \\
\bottomrule
\end{tabu}
\caption{160M Parameter Model Results}
\end{subtable}

\begin{subtable}{\textwidth}
\centering
\begin{tabu}{l|ccc|lll|lll} 
\toprule
\multicolumn{1}{c|}{\multirow{2}{*}{\textbf{Model}}}                                                    & \multicolumn{3}{c|}{\textbf{Consistency ($\uparrow$)}}                                                       & \multicolumn{3}{c|}{\textbf{Case Error ($\downarrow$)}}                                                    & \multicolumn{3}{c}{\textbf{Inject Error ($\downarrow$)}}                                                    \\
                                                                                 & \textbf{He}                        & \textbf{She}                       & \textbf{Xe}                        & \multicolumn{1}{c}{\textbf{He}}   & \multicolumn{1}{c}{\textbf{She}}  & \multicolumn{1}{c|}{\textbf{Xe}}   & \multicolumn{1}{c}{\textbf{He}}   & \multicolumn{1}{c}{\textbf{She}}  & \multicolumn{1}{c}{\textbf{Xe}}     \\ 
\midrule
\rowfont{\color{gray}}\textsc{$\text{T}_{\text{Orig}} + \text{M}_{\text{Base}}$} & 72.82\textsubscript{1.97}          & \textbf{55.85\textsubscript{2.21}} & 0.05\textsubscript{0.08}           & \textbf{2.87\textsubscript{0.74}} & \textbf{7.90\textsubscript{1.18}} & 79.90\textsubscript{1.79}          & \textbf{4.15\textsubscript{0.90}} & \textbf{3.49\textsubscript{0.85}} & 89.85\textsubscript{1.33}           \\
\textsc{$\text{T}_{\text{Orig}} + \text{M}_{\text{Full}}$}                       & \textbf{79.03\textsubscript{1.82}} & 42.10\textsubscript{2.21}          & 18.36\textsubscript{1.74}          & 9.28\textsubscript{1.31}          & 19.69\textsubscript{1.74}         & 39.54\textsubscript{2.18}          & 12.82\textsubscript{1.49}         & 19.79\textsubscript{1.77}         & 56.62\textsubscript{2.21}           \\
\textsc{$\text{T}_{\text{Orig}} + \text{M}_{\text{Lex}}$}                        & 69.85\textsubscript{2.03}          & 42.10\textsubscript{2.23}          & 19.85\textsubscript{1.74}          & 11.85\textsubscript{1.41}         & 20.87\textsubscript{1.79}         & 48.10\textsubscript{2.23}          & 16.97\textsubscript{1.67}         & 11.79\textsubscript{1.46}         & 54.51\textsubscript{2.18}           \\
\textsc{$\text{T}_{\text{PTP}} + \text{M}_{\text{Full}}$}                        & 63.64\textsubscript{2.13}          & 48.21\textsubscript{2.23}          & \textbf{56.21\textsubscript{2.21}} & 6.77\textsubscript{1.10}          & 14.51\textsubscript{1.59}         & 31.69\textsubscript{2.05}          & 14.36\textsubscript{1.54}         & 14.97\textsubscript{1.56}         & 20.67\textsubscript{1.79}           \\
\textsc{$\text{T}_{\text{PTP-B}} + \text{M}_{\text{Full}}$}                      & 82.31\textsubscript{1.69}          & 48.82\textsubscript{2.26}          & 19.03\textsubscript{1.72}          & 6.36\textsubscript{1.08}          & 16.82\textsubscript{1.67}         & 31.03\textsubscript{2.05}          & 11.54\textsubscript{1.41}         & 17.38\textsubscript{1.67}         & 26.21\textsubscript{1.95}           \\
\textsc{$\text{T}_{\text{PTP}} + \text{M}_{\text{Lex}}$}                         & 68.31\textsubscript{2.05}          & 45.74\textsubscript{2.26}          & 40.62\textsubscript{2.15}          & 9.95\textsubscript{1.36}          & 23.69\textsubscript{1.90}         & 30.87\textsubscript{2.03}          & 16.72\textsubscript{1.67}         & 13.38\textsubscript{1.54}         & 13.59\textsubscript{1.51}           \\
\textsc{$\text{T}_{\text{PTP-B}} + \text{M}_{\text{Lex}}$}                       & 69.44\textsubscript{2.03}          & 35.59\textsubscript{2.13}          & 49.18\textsubscript{2.23}          & 9.28\textsubscript{1.28}          & 24.31\textsubscript{1.92}         & \textbf{28.72\textsubscript{2.00}} & 17.74\textsubscript{1.67}         & 17.44\textsubscript{1.67}         & \textbf{12.26\textsubscript{1.46}}  \\
\bottomrule
\end{tabu}
\caption{410m Parameter Model Results}
\end{subtable}

\begin{subtable}{\textwidth}
\centering
\begin{tabu}{l|ccc|lll|lll} 
\toprule
\multicolumn{1}{c|}{\multirow{2}{*}{\textbf{Model}}}                                                    & \multicolumn{3}{c|}{\textbf{Consistency ($\uparrow$)}}                                                       & \multicolumn{3}{c|}{\textbf{Case Error ($\downarrow$)}}                                                    & \multicolumn{3}{c}{\textbf{Inject Error ($\downarrow$)}}                                                    \\
                                                                                 & \textbf{He}                        & \textbf{She}                       & \textbf{Xe}                        & \multicolumn{1}{c}{\textbf{He}}   & \multicolumn{1}{c}{\textbf{She}}  & \multicolumn{1}{c|}{\textbf{Xe}}   & \multicolumn{1}{c}{\textbf{He}}   & \multicolumn{1}{c}{\textbf{She}}  & \multicolumn{1}{c}{\textbf{Xe}}     \\ 
\midrule
\rowfont{\color{gray}}\textsc{$\text{T}_{\text{Orig}} + \text{M}_{\text{Base}}$} & 78.46\textsubscript{1.82}          & \textbf{66.56\textsubscript{2.08}} & 0.26\textsubscript{0.23}           & 3.54\textsubscript{0.85}          & \textbf{3.03\textsubscript{0.77}} & 76.00\textsubscript{1.90}          & \textbf{3.69\textsubscript{0.85}} & \textbf{3.44\textsubscript{0.79}} & 92.77\textsubscript{1.15}           \\
\textsc{$\text{T}_{\text{Orig}} + \text{M}_{\text{Full}}$}                       & 76.72\textsubscript{1.87}          & 58.72\textsubscript{2.18}          & 17.90\textsubscript{1.72}          & 8.10\textsubscript{1.23}          & 25.18\textsubscript{1.92}         & 36.46\textsubscript{2.15}          & 24.72\textsubscript{1.90}         & 24.56\textsubscript{1.92}         & 36.31\textsubscript{2.13}           \\
\textsc{$\text{T}_{\text{Orig}} + \text{M}_{\text{Lex}}$}                        & 80.05\textsubscript{1.77}          & 56.00\textsubscript{2.18}          & 15.49\textsubscript{1.59}          & 5.64\textsubscript{1.03}          & 17.90\textsubscript{1.72}         & 40.82\textsubscript{2.21}          & 16.62\textsubscript{1.69}         & 35.49\textsubscript{2.13}         & 55.23\textsubscript{2.21}           \\
\textsc{$\text{T}_{\text{PTP}} + \text{M}_{\text{Full}}$}                        & \textbf{84.72\textsubscript{1.64}} & 56.46\textsubscript{2.21}          & \textbf{44.97\textsubscript{2.21}} & 4.56\textsubscript{0.92}          & 20.31\textsubscript{1.79}         & 44.92\textsubscript{2.21}          & 24.10\textsubscript{1.90}         & 18.00\textsubscript{1.69}         & 20.05\textsubscript{1.77}           \\
\textsc{$\text{T}_{\text{PTP-B}} + \text{M}_{\text{Full}}$}                      & 71.90\textsubscript{2.00}          & 53.95\textsubscript{2.21}          & 35.69\textsubscript{2.13}          & 8.41\textsubscript{1.26}          & 18.00\textsubscript{1.72}         & 40.10\textsubscript{2.18}          & 19.13\textsubscript{1.79}         & 22.31\textsubscript{1.82}         & \textbf{18.51\textsubscript{1.72}}  \\
\textsc{$\text{T}_{\text{PTP}} + \text{M}_{\text{Lex}}$}                         & 76.77\textsubscript{1.90}          & 44.62\textsubscript{2.21}          & 36.26\textsubscript{2.10}          & \textbf{2.56\textsubscript{0.69}} & 18.62\textsubscript{1.72}         & \textbf{31.54\textsubscript{2.08}} & 12.05\textsubscript{1.44}         & 24.56\textsubscript{1.90}         & 19.33\textsubscript{1.74}           \\
\textsc{$\text{T}_{\text{PTP-B}} + \text{M}_{\text{Lex}}$}                       & 79.74\textsubscript{1.82}          & 57.85\textsubscript{2.18}          & 35.64\textsubscript{2.13}          & 4.67\textsubscript{0.92}          & 14.62\textsubscript{1.56}         & 33.13\textsubscript{2.10}          & 20.10\textsubscript{1.77}         & 26.72\textsubscript{1.97}         & 27.23\textsubscript{1.97}           \\
\bottomrule
\end{tabu}
\caption{1.4B Parameter Model Results}
\end{subtable}
\caption{Model Size Comparisons at Data Split=10}
\label{tbl:appendix_size_compare}
\end{table*}

\autoref{app:70m_compare} provides results across all data splits for the 70M model. \autoref{tbl:appendix_size_compare} provides results across model sizes for the 10\% data resource ablation, so as to best mimic real-world low-resource circumstances.


\section{Ablations Across Size and Data Resource}
\label{sec:appendix_size_compare}

\end{document}